\documentclass[twoside,leqno,twocolumn]{article}

\usepackage[letterpaper]{geometry}

\usepackage{ltexpprt}

\usepackage{cite}
\usepackage{amsmath,amssymb,amsfonts}
\usepackage{algorithmic}
\usepackage{graphicx}
\usepackage{textcomp}
\usepackage{xcolor}

\usepackage[utf8]{inputenc} 
\usepackage[T1]{fontenc}    
\usepackage[backref=page]{hyperref}       
\usepackage{url}            
\usepackage{booktabs}       
\usepackage{nicefrac}       
\usepackage{microtype}      
\usepackage{multicol}
\usepackage{lipsum}
\usepackage{mwe}
\usepackage[english]{babel}
\usepackage[utf8]{inputenc}

\usepackage[nottoc]{tocbibind}
\usepackage{bbm}
\usepackage[thinlines]{easytable}
\usepackage{svg}
\usepackage{subcaption}
\usepackage{algorithm}

\def\BibTeX{{\rm B\kern-.05em{\sc i\kern-.025em b}\kern-.08em
    T\kern-.1667em\lower.7ex\hbox{E}\kern-.125emX}}

\begin{document}

\newcommand\relatedversion{}
\setlength{\belowcaptionskip}{-10pt}

\title{\Large Scalable Batch Acquisition for Deep Bayesian Active Learning \relatedversion}
\author{Aleksandr Rubashevskii\thanks{Skolkovo Institute of Science and Technology, Moscow, Russia. \{aleksandr.rubashevskii, d.kotova\}@skoltech.ru}
\and Daria Kotova\footnotemark[1]
\and Maxim Panov\thanks{Technology Innovation Institute, Abu Dhabi, UAE. maxim.panov@tii.ae}}



\date{}

\maketitle


\fancyfoot[R]{}





\begin{abstract} \small\baselineskip=9pt 
    In deep active learning, it is especially important to choose multiple examples to markup at each step to work efficiently, especially on large datasets.
    At the same time, existing solutions to this problem in the Bayesian setup, such as BatchBALD, have significant limitations in selecting a large number of examples, associated with the exponential complexity of computing mutual information for joint random variables.
    We, therefore, present the Large BatchBALD algorithm, which gives a well-grounded approximation to the BatchBALD method that aims to achieve comparable quality while being more computationally efficient.
    We provide a complexity analysis of the algorithm, showing a reduction in computation time, especially for large batches.
    Furthermore, we present an extensive set of experimental results on image and text data, both on toy datasets and larger ones such as CIFAR-100.
\end{abstract}


\section{Introduction}
\label{sec:intro}
    In supervised machine learning tasks, the quality and volume of the training data play essential roles in achieving high performance.
    However, the process of data collection and labeling is often expensive, requiring a huge amount of time and resources~\cite{roh2021survey,paullada2021survey}.
    Therefore, active learning (AL) techniques, that choose the most informative samples for model training, minimizing the collection and annotation budget, are crucial in practice~\cite{ren2021survey}.
    Active learning methods are successfully applied to the various types of data: tabular~\cite{tsymbalov2018dropout}, image~\cite{gal2017al}, text~\cite{shelmanov2019active}, audio~\cite{riccardi2005alaudio}, video~\cite{vondrick2011alvideo} and others. 
    Especially, AL can be helpful in the case of real-world data which requires involvement of subject-matter experts, namely medical~\cite{budd2021survey} and manufacturing data~\cite{lv2006deep}.
    
    In this work, we consider a pool-based active learning problem for classification tasks.
    It is assumed that there is a small amount of labeled data and a big unlabeled pool to select an object for annotation from.
    Selection procedure is carried out according to a certain criterion that is usually based on a so-called acquisition function.
    Acquisition function is maximized over the most informative samples in terms of model uncertainty measure or expected error.
    For example, as an acquisition function one uses variance reduction~\cite{hoi2006batch}, entropy~\cite{paola2000entropy} or mutual information (also known as BALD)~\cite{houlsby2011bald} maximization and others.
    Then, selected samples are labeled and added to the already labeled dataset for further training.
    
    At each step of the active learning cycle, one or several pool samples can be selected for annotation.
    By selecting one sample to label at each step, one can greedily assemble an optimal set of labeled data for training.
    However, with a large number of objects in the dataset, this strategy becomes inefficient as frequent model retraining becomes very computationally expensive. 
    In this case, it is better to select multiple objects from the pool at each active learning step.
    
    However, an acquisition of multiple objects at a time, leads to the problem of selecting similar training samples and data redundancy.
    Indeed, the majority of existing acquisition function (for example, BALD) do not take into account the interaction of samples within a batch. That's why very similar samples can be selected and the resulting active-learned training dataset can have many redundant points. It leads to model performance degradation and excessive use of resources. 
    One of the state-of-the-art methods that partially copes with this problem is an extension of the BALD method, namely BatchBALD~\cite{gal2019batchbald}.
    Its idea is to calculate mutual information in a batch manner using multiple network outputs as a joint random variable.
    This approach allows one to account for interactions between samples in a batch manner, preventing the selection of similar samples, but greatly increasing computational time and complexity.
    
    In this work, we aim to improve over BatchBALD by dealing with its main weaknesses, namely its high computational complexity.
    The main contributions of this work can be summarized as follows.
    \begin{itemize}
        \item We propose a new active learning algorithm called \textit{Large BatchBALD}, which is an approximation of the BatchBALD method that allows to improve computational efficiency, while keeping the diversity of samples in a batch. The details can be found in Section~\ref{sec:methodology}.

        \item We analyze the complexity of the proposed algorithm (see Section~\ref{sec:complexity}), showing a reduction in running time compared to the original BatchBALD, especially for large batches.

        \item We provide an extensive experimental study showing the improved efficiency and successful performance of Large BatchBALD in active learning for image and text classification tasks compared to state-of-the-art approaches, see Section~\ref{sec:experiments}.

        \item We additionally study how the stochastic sampling can help to further improve the results compared to the greedy approaches, see Sections~\ref{sec:stochastic_acquisition} and~\ref{sec:experiments}.
    \end{itemize}
    
    We would like to note that a similar idea was also introduced in the blog-post~\footnote{\url{https://web.archive.org/web/20220702232856/https://blog.blackhc.net/2022/07/kbald/}} and then as a paper~\cite{kirsch2023speeding}.
    Despite the similarity of approaches, both ideas were developed concurrently and independently of each other.

    %


\section{Methodology}
\label{sec:methodology}
\subsection{Problem setting}
    In this paper, we consider a pool-based active learning problem statement. 
    This implies that we have a small amount of labeled examples $D_{\text{train}} = \{ x_i, y_i\}_{i=1}^{N} $, $y_i \in \{1, \ldots, C\}$ and a much larger pool of unlabeled data $D_{\text{pool}} = \{x_j\}_{j=1}^{N_{\text{pool}}}$. We consider a Bayesian model $M$ with parameters $\theta \sim p(\theta \mid D_{\text{train}})$. Here, conditioning on $D_{\text{train}}$ emphasizes that the model was trained using this dataset.
    Acquired samples from $D_{\text{pool}}$ are selected as the ones that maximize a so-called acquisition function $A$:
    \begin{equation}
        x^* = \arg\max_{x \in D_{\text{pool}}}~A(x \mid M, D_{\text{train}}).
    \label{eq:max_acq_func}
    \end{equation}
    It maps each example from the unlabeled pool to a numerical value using some information criteria or uncertainty measure.
    Acquired samples from the pool are selected for oracle labeling, and then these examples are added to the existing labeled data $D_{\text{train}}$.
    The target model is then trained on the currently labeled amount of data.
    This procedure is repeated throughout the active learning cycle and continues until the total budget of the algorithm is exhausted.
    At each step of the cycle, the acquisition function is recalculated.
    The quality of the model is evaluated on the test data $D_{\text{test}}$ and compared at each step of the active learning procedure.
    

    There are various acquisition functions applicable in AL. One of the standard ones is the Least Confident score:
    \begin{equation}
        A(x) = 1 - \max_{c \in C} p(y = c \mid x, \theta),
    \label{eq:least_conf_acq_func}
    \end{equation}
    where $\theta$ are model parameters.
    That is, we choose a sample for which the model is most uncertain about the class predicted.
    
    In this work, we use both MC-dropout and deep ensembles approaches for more accurate capturing of uncertainty for deep neural networks.
    In the case of deep ensembles, the models of the same architecture but trained from different initializations form an ensemble. In the case of MC-dropout, the dropout sampling on inference is used to obtain a set of networks with different parametrization using different dropout masks.
    Formally, in both cases the final prediction can be written as
    \begin{equation}
        \bar{p}_k(y = c \mid x, D_{\text{train}}) = \dfrac{1}{k} \sum_{i = 1}^k p(y = c \mid x, \theta_i),
    \label{eq:mc_deep_ens}
    \end{equation}
    where $\theta_i$ is the model parameters for the $i$-th model and $k$ is the number of models in a set.
    It can be either the number of network initializations in the case of deep ensemble, or the number of forward passes in the case of MC-dropout. Both MC-dropout and ensembling can be seen as instances of the general Bayesian formulation, with model parameters being samples from the posterior distribution: $\theta_i \sim p(\theta \mid D_{\text{train}}), ~ i = 1, \dots, k$.

    While one can directly use the averaged predictive distribution~\eqref{eq:mc_deep_ens} in the least confident acquisition function~\eqref{eq:least_conf_acq_func}, it might be beneficial to extract some additional information from the posterior on top of the predictive mean. Next, we will focus on the family of entropy-based acquisition functions that allow to achieve that.

\subsection{Entropy-based acquisition functions}
\subsubsection{Entropy}
    The entropy maximization criterion is also one of the basic ways of selecting examples for AL:
    \begin{align*}
        & \mathrm{H}[y \mid x, D_{\text{train}}] \\
        & = -\sum_{c = 1}^C \bar{p}_k(y = c \mid x, D_{\text{train}})
        \cdot \log \bar{p}_k(y = c \mid x, D_{\text{train}}).
    \end{align*}
    %
    It again uses the predictive mean only by looking at the entropy of this distribution. This criterion is maximized for samples having similar probabilities predicted for the different classes.

\subsubsection{Bayesian Active Learning by Disagreement (BALD)}
    Starting from the entropy, one can construct criteria that look at the disagreement between the models in the Bayesian framework. The original BALD criterion~\cite{houlsby2011bald} is formulated as the conditional mutual information $I(\theta, y \mid x, D_{\text{train}})$ between unknown (unobserved) output $y$ and latent parameters $\theta$, conditioned on input variable $x$ and observed data $D_{\text{train}}$.
    Note, that the BALD criterion can be written in the $y$-space of mutual information and expressed as follows:
    \begin{align*}
      I_{\text{BALD}}(y; \theta) = \mathrm{H}[y \mid x, D_{\text{train}}] - \mathbb{E}_{\theta \sim p(\theta \mid D_{\text{train}})} \bigl[\mathrm{H}[y \mid x, \theta] \bigr], 
    \label{eq:bald_paper}
    \end{align*}
    where $\mathrm{H}[y \mid x, D_{\text{train}}]$ is an entropy of model output $y$ conditioned on data sample $x$ and train data, $\mathrm{H}[y \mid x, \theta]$ is an entropy of model output $y$ conditioned on data sample $x$ and sampled latent model parameters $\theta \sim p(\theta \mid D_{\text{train}})$ which are integrated out by the expectation $\mathbb{E}_{\theta \sim p(\theta \mid D_{\text{train}})}$. 
    In other words, it calculates the difference between the entropy of marginal predictive distribution and posterior mean conditional entropy.
    BALD intuition is that it seeks for data samples in whose outputs $y$ the model is the most uncertain (leads to high marginal entropy), while being certain about individual model parameters (leads to confident predictions but highly diverse).
    In general, a continuous case BALD acquisition function is expressed as a KL divergence between $p(y; \theta)$ and $p(y) p(\theta)$:
    \begin{equation}
      I(y; \theta) = \int_{\theta} \int_{y} p(y; \theta) \log \dfrac{p(y; \theta)}{p(y) p(\theta)} \mathrm{d} y \mathrm{d} \theta.
    \label{eq:bald_integr}
    \end{equation}
    In terms of batch active learning, when an acquisition batch where $y_{1:b} := y_1, \ldots, y_b$ consists of $b > 1$ data samples, the BALD score is the sum of individual scores for each of $b$ objects in the batch:
    \begin{equation}
      I_{\text{BALD}}(y_{1:b}; \theta) = \sum_{i=1}^b I(y_i; \theta).
    \label{eq:bald_batch}
    \end{equation}
    This approach has a serious drawback, namely, it does not take into account pairwise interactions of the data samples in batches. As a result, BALD tends to acquire a batch of similar examples, leading to suboptimal performance.

\subsubsection{BatchBALD}
    To diversify samples in a batch, the BatchBALD acquisition function was proposed~\cite{gal2019batchbald}. It is formulated as mutual information between a batch of observations and latent parameters:
    \begin{align*}
        I_{\text{BB}}(y_{1:b}; \theta) &= \mathrm{H}[y_1, \ldots, y_b \mid x_1, \ldots, x_b, D_{\text{train}}] \\
        &- \mathbb{E}_{\theta \sim p(\theta \mid D)} \bigl[\mathrm{H}[y_1, \ldots, y_b \mid x_1, \ldots, x_b, \theta] \bigr],
    \label{eq:batchbald_paper}
    \end{align*}
    where $y_{1:b} := y_1, \ldots, y_b$ is treated as a joint random variable, $b$ is the batch acquisition size.
    BatchBALD calculates mutual information between model output and model parameters but in a batch sense, that is, considering inter-variable correlation and taking a batch of outputs as a joint random variable.
    It allows to account for variable interconnections and provides diverse data sampling.
    In the continuous case, it is again given by the corresponding KL divergence: 
    \begin{equation}
      I(y_{1:b}; \theta) = \int_{\theta} \int_{y_{1:b}} p(y_{1:b}; \theta) \log \dfrac{p(y_{1:b}; \theta)}{p(y_{1:b}) p(\theta)} \mathrm{d} y_{1:b} \mathrm{d} \theta. \hspace{-8pt}
    \label{eq:batchbald_integr}
    \end{equation}
    While accounting nicely for the correlation between observations, BatchBALD criterion is often computationally expensive, especially for large batches (see Section~\ref{sec:complexity} for the complexity analysis). In the next section, we are going to provide its more computationally feasible alternative.

  \begin{table*}[t!]
    \centering
    \begin{tabular}{|c|c|c|}
      \hline
      BALD & BatchBALD & Large BatchBALD  \\ \hline
      $\mathcal{O}((b + k) \cdot |D_\text{pool}|)$ & $\mathcal{O}(b \cdot c \cdot \min\{c^b, m\} \cdot |D_\text{pool}| \cdot k)$ & $\mathcal{O}(|D_\text{pool}|^2 \cdot k \cdot c + |D_\text{pool}| \cdot (b + k))$ \\ \hline
    \end{tabular}
    \caption{Complexity of BALD-based algorithms, where $c$ is the number of classes, $b$ is the batch acquisition size, $k$ is the number of MC-dropout samples, $m$ is the number of MC-sampled output configurations $y_{1: b-1}$.}
    \label{tab:alg_comp}
  \end{table*}

\subsection{Large BatchBALD}
    One of the possible generalizations of mutual information is total correlation~\cite{watanabe1960totcorr} between $b$ random variables, which is defined as:
    \begin{equation}
      C(y_{1:b}) = \int \limits_{y_1:b} p(y_{1:b}) \log \dfrac{p(y_{1:b})}{\prod_{i = 1}^{b} p(y_i)} \mathrm{d} y_{1:b}.
    \label{eq:tot_corr_def}
    \end{equation}
    It measures inter-variable dependencies, always positive and is nullified if and only if all the variables are independent of each other.
    Note that its form doesn't include model latent parameters $\theta$.

    The main idea of introducing of total correlation in this work is that it is exactly equal to the difference between BALD and BatchBALD acquisition functions:
    \begin{equation}
      \sum_{i=1}^b I(y_i; \theta) - I_{\text{BB}}(y_{1:b}; \theta) - C(y_{1:b}) = 0,
    \label{eq:b_bb_lbb}
    \end{equation}
    where $C(y_{1:b}) = C(y_1; y_2; \ldots; y_b)$ is the mutual information of $b$ random variables (i.~e., generalization of mutual information of two variables), $b$ is an acquisition batch size. A complete derivation of equation~\eqref{eq:b_bb_lbb} can be found in Supplementary Material, Section~\ref{sec:proof}.
    
    Calculation of total correlation and, therefore, calculation of BatchBALD can be extremely time-consuming. However, one can overcome this issue inspiring from another possible form of total correlation~\cite{srinivasa2005review} that uses mutual information of all possible variable subscripts:
    \begin{equation}
      \begin{aligned}
        C(y_{1:b}) 
        &= \sum_{i \neq j}^b I\left(y_{i} ; y_{j}\right) - \sum_{i \neq j \neq k}^b I\left(y_{i} ; y_{j} ; y_{k}\right) \\ 
        &+\ldots + (-1)^b I\left(y_{1} ; y_{2} ; \ldots ; y_{b}\right).
      \end{aligned}
    \label{eq:tot_corr_alter}
    \end{equation}
    The idea is to neglect higher order terms and use total correlation approximation with only pairwise mutual information components:
    \begin{equation}
      \hat{C}(y_{1:b}) = \sum \limits_{i=1}^b \sum \limits_{j \neq i}^b I\left(y_{i} ; y_{j}\right).
    \label{eq:tot_corr_approx}
    \end{equation}
    %
    Using the approximation of total correlation, we obtain
    \begin{equation}
        I_{\text{BB}}(y_{1:b}; \theta) 
        \approx \sum_{i=1}^b I(y_i, \theta) - \hat{C}(y_{1:b}).
    \label{eq:bb_approx}
    \end{equation}
    %
    We denote this approximation as Large BatchBALD (LBB), and the resulting formula for it becomes
    \begin{equation}
      I_{\text{LBB}}(y_{1:b}; \theta) := \sum \limits_{i=1}^b I(y_i, \theta) - \sum \limits_{i=1}^b \sum \limits_{j \neq i}^b I(y_i; y_j).
    \end{equation}
    Importantly, LBB is significantly less computationally expensive compared to BatchBALD, see the complexity analysis in Section~\ref{sec:complexity}.

\subsection{Computational complexity}
\label{sec:complexity}
  In this section, we discuss the computational complexity for the BALD, BatchBALD, and Large BatchBALD algorithms, see Table~\ref{tab:alg_comp}.
  We also give some intuition how LBB, using BALD and pairwise mutual information, can significantly improve the complexity of the BatchBALD algorithm, especially noticeable in the case of large batches.

\subsubsection{BALD}
  BALD time complexity is $\mathcal{O}((b + k) \cdot |D_\text{pool}|)$ and consists of calculating for each element of $D_\text{pool}$ 2 components: $\mathcal{O} (b \cdot |D_\text{pool}|)$ is a cost to compute predictive distribution and $\mathcal{O} (k \cdot |D_\text{pool}|)$ is a cost to compute entropies in output space.

\subsubsection{BatchBALD}
  To compute the exact joint entropies, we have to compute all possible configurations of the $p(y_1, \ldots, y_b)$ and evaluate by averaging over $p(y_1, \ldots, y_b \mid \theta)$.
  To compute approximate joint entropies, we have to sample possible configurations of the $y_i$ from $p(y_1, \ldots, y_b)$ stratified by $p(\theta)$ and evaluate $p(y_1, \ldots, y_b)$ by averaging over $p(y_1, \ldots, y_b \mid \theta)$.

  BatchBALD complexity is $\mathcal{O}(b \cdot c \cdot \min\{c^b, m\} \cdot |D_\text{pool}| \cdot k)$, where $c$ is the number of classes, $k$ is the number of MC-dropout samples, and $m$ is the number of MC-sampled configurations of $y_{1: b-1}$, $|D_\text{pool}|$ is a volume of unlabeled pool data.
  A description of this result and corresponding details can be found in Supplementary Material, Section~\ref{sec:appendix_complexity}.

\subsubsection{Large BatchBALD}
  Large BatchBALD time complexity is equal to the sum of BALD complexity and total correlation approximation complexity.
  BALD complexity, as noted above, is $\mathcal{O}((b + k) \cdot |D_\text{pool}|)$.
  Total correlation approximation complexity is $\mathcal{O} \left(2 \cdot \dfrac{|D_\text{pool}| \cdot |D_\text{pool} - 1|}{2} \cdot k \cdot c \right) = \mathcal{O} \left(|D_\text{pool}|^2 \cdot k \cdot c \right)$, see Supplementary Material, Section~\ref{sec:mut_info_calc} for details.
  It means, that Large BatchBALD complexity is $\mathcal{O}(|D_\text{pool}|^2 \cdot k \cdot c + |D_\text{pool}| \cdot (b + k))$.
    
  The resulting complexity has the following intuition behind.
  The given asymptotics denotes a linear dependence on the size of the batch, as in BALD.
  Thus, Large BatchBALD scales to the size of a batch consisting of hundreds of elements without significant costs.
  At the same time, the original BatchBALD algorithm works in a reasonable time only with batches consisting of tens of elements due to the calculation of mutual information of a joint random variable.
  In general, adding large batches in an active learning problem is a common practical scenario.
  With a huge pool of unlabeled data, adding a small amount of data up to a few tens will have little effect on the performance of the final model.
  Thus, it is computationally more efficient to be able to acquire large batches for annotation and training.
  Furthermore, LBB works equally well with batches of tens of elements and already shows computational superiority in comparison with BatchBALD, see Table~\ref{tab:ens_mnist_time_comparison}.

\subsection{Beyond greedy approaches via sampling}
\label{sec:stochastic_acquisition}
    While BatchBALD and its modifications are efficient in obtaining diverse batches, one can propose alternative strategies to achieve a similar effect. One natural way is to step aside from greedy sampling and introduce stochasticity into the procedure.
    The idea is to convert the resulting scores to a distribution and then sample from it.
    In this case, we raise the scores to some power $\alpha > 0$ and normalize the resulting values.
    Thus, the probability of selecting a sample $x$ with an acquisition function equal to $A(x)$ from the unlabeled pool $D_\text{pool}$ is
    \begin{equation}
      p_{\alpha}(x) = \dfrac{A^\alpha(x)}{\sum \limits_{x_i \in D_\text{pool}} A^\alpha(x_i)}.
    \label{eq:power_def}
    \end{equation}
    In this work we consider such extension for the LBB and BALD algorithms, and call them Power Large BatchBALD (PLBB) and PowerBALD (PBALD)~\cite{kirsch2021stoch_batch}. Thus, in the case of PLBB, the acquisition function is $A(x) = I_{\text{LBB}}(y; \theta)$, and in the case of PBALD it is $A(x) = I_{\text{BALD}}(y; \theta)$. Here $y$ is the model output on the sample $x$ and $\theta$ is the vector of model parameters.
    The magnitude of the power $\alpha$ in this case determines how much of a stochastic effect is present: with a smaller power the random effect is greater, with a greater power examples with larger scores are even more likely to be taken, and the random effect appears less.


\section{Experiments}
\label{sec:experiments}    
    For all datasets, the initial training set is balanced by the number of samples of each class.
    All datasets with the repetition option use each incoming object more than once ($4$ in our case) with a small Gaussian noise applied.
    
    After each addition of new samples to the training set, the network is trained from scratch.
    All models use Glorot initialization.
    The parameters of MC-dropout experiments are similar to the work~\cite{gal2019batchbald} settings, deep ensembles experiments are performed with an ensemble of $5$ models.
    
    We measure the accuracy of the model prediction on a test dataset, depending on the amount of training data obtained by different algorithms.
    All results are obtained as the average of the $5$ runs, and the corresponding standard deviation is shown as filled error bars.

\subsection{MNIST and its variations}
\subsubsection{Experimental setup}
    The first group of experiments deals with MNIST and its extensions: MNIST~\cite{lecun1998mnist}, Repeated MNIST (RMNIST; \cite{gal2019batchbald}), and Fashion MNIST (FMNIST; \cite{xiao2017fmnist}).
    MNIST is a standard machine learning dataset suitable for active learning that consists of handwritten digit images, including $60,000$ images from $10$ classes.
    RMNIST is an extension of the MNIST dataset in which each image is repeated several times with a small Gaussian noise applied.
    FMNIST is a fashion product dataset containing $70,000$ images of $10$ classes.

    Model architecture for MNIST, RMNIST and FMNIST datasets is taken similar as in~\cite{gal2019batchbald} for the MC-dropout uncertainty case.
    For experiments with deep ensembles, the same architecture was adapted to use multiple initializations of the same network to form an ensemble.
    Note that ensembles are more time-consuming than MC-dropout.
    While MC-dropout requires multiple forward passes on inference of the same network, for ensembles one needs to fully train multiple networks.
    Nevertheless, when using deep ensembles, better model quality can be achieved by better calibration of the resulting models~\cite{beluch2018ensembles}.
    We compared the performance of the AL algorithms for both ensembles and MC-dropout uncertainty estimates on acquisition batches of $10$ and $20$ images on the specified datasets.
    
    \begin{figure*}[htb]
      \begin{subfigure}{0.32\textwidth}
        \includegraphics[width=\linewidth]{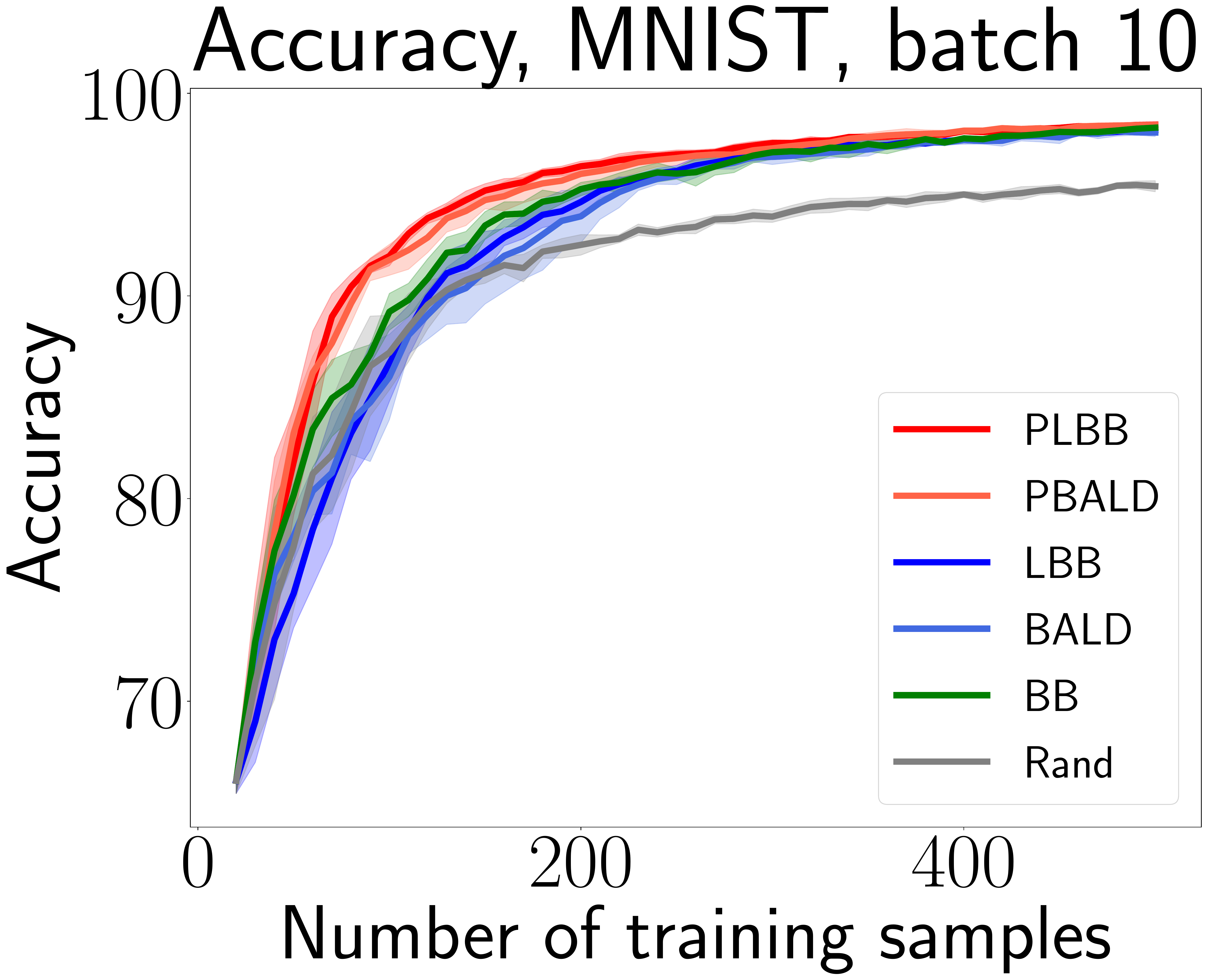}
        \caption{} \label{fig:ens_mnist_batch10}
      \end{subfigure}
      \hspace*{\fill}
      \begin{subfigure}{0.32\textwidth}
        \includegraphics[width=\linewidth]{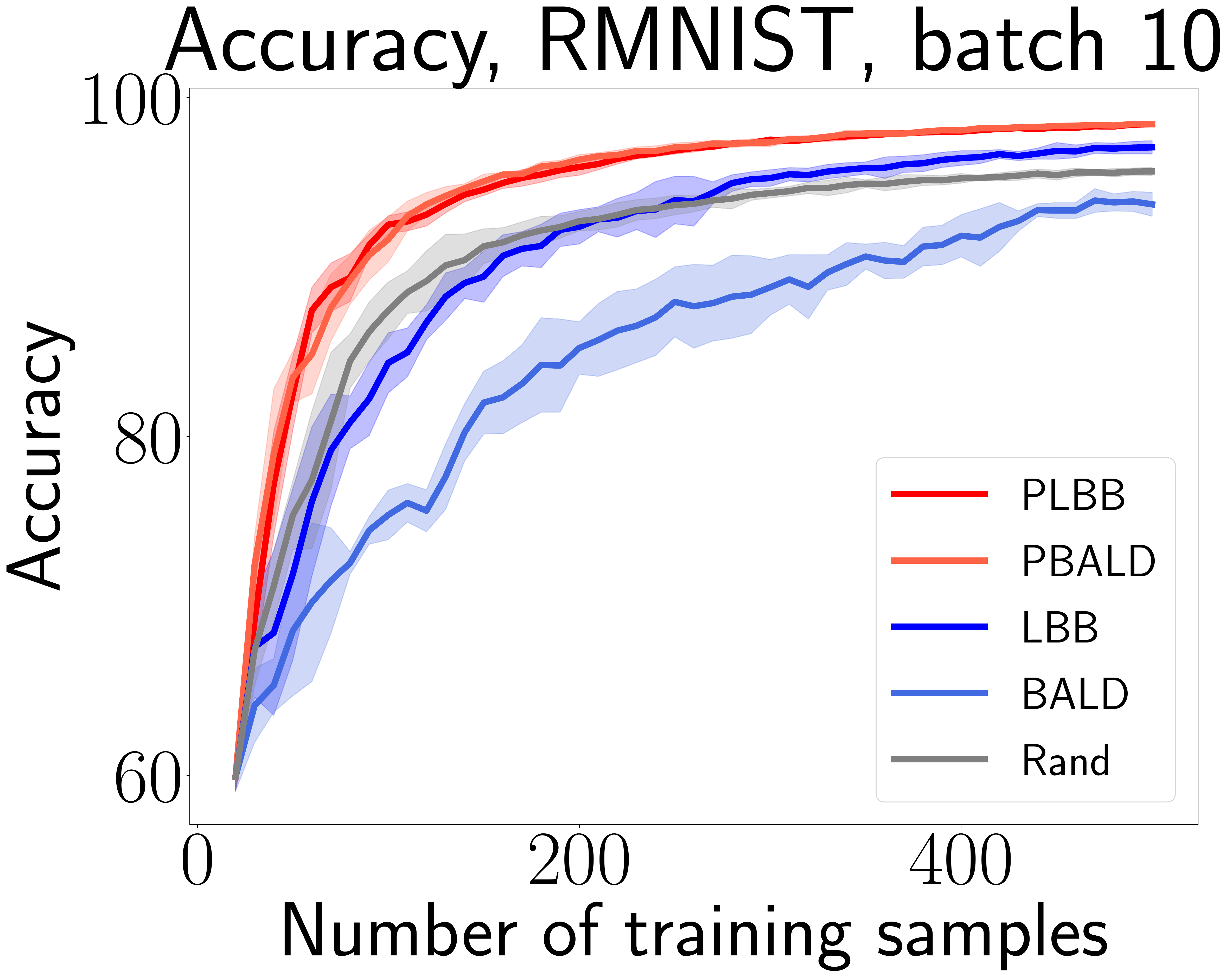}
        \caption{} \label{fig:ens_rmnist_batch10}
      \end{subfigure}%
      \hspace*{\fill}
      \begin{subfigure}{0.32\textwidth}
          \includegraphics[width=\linewidth]{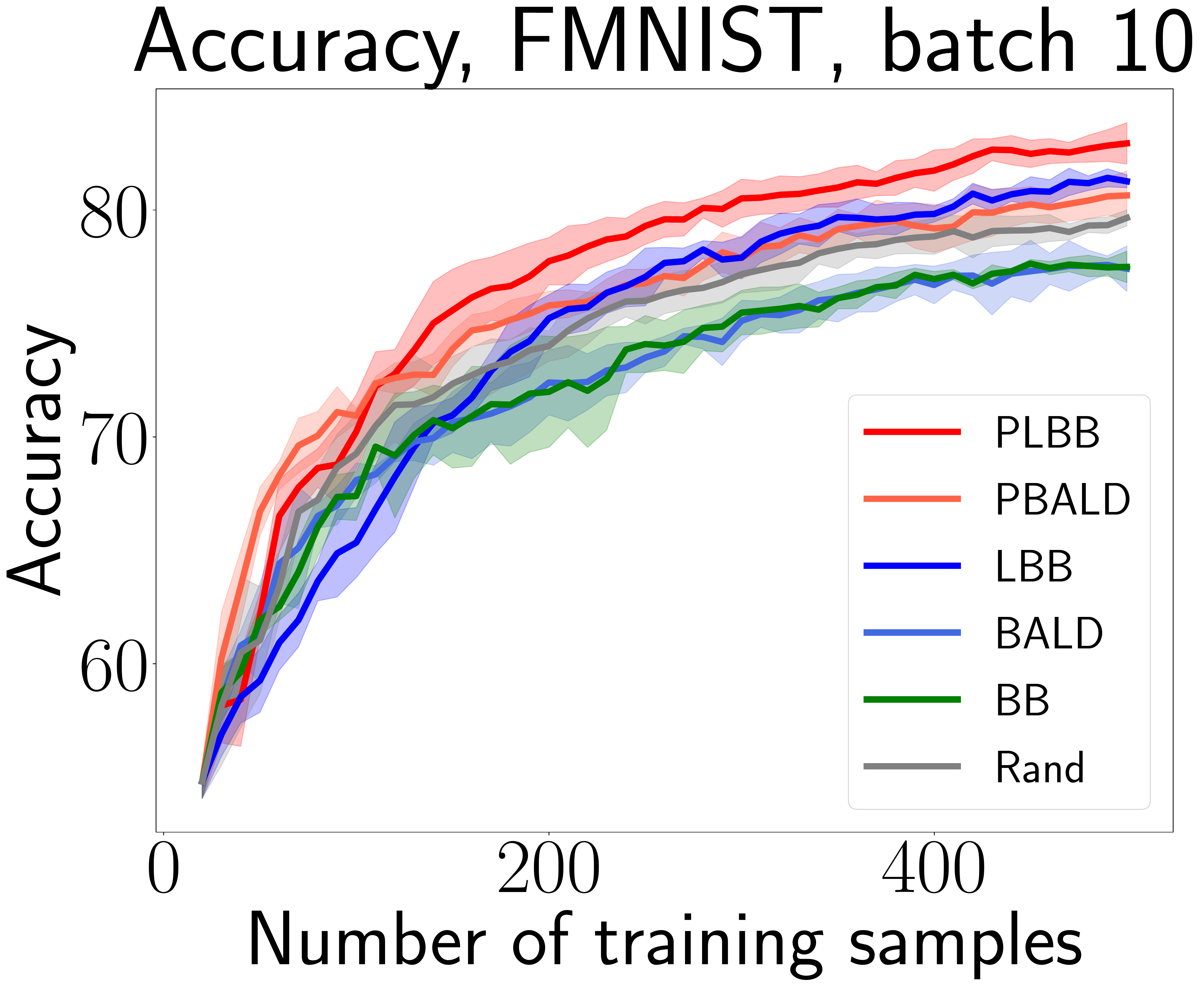}
        \caption{} \label{fig:ens_fmnist_batch10}
      \end{subfigure}
    
      \caption{Performance comparison of AL algorithms on deep ensembles. Datasets: (a)~MNIST. (b)~RMNIST. (c)~FMNIST. LBB shows better performance than BALD, and their randomized extensions, PLBB and PBALD, lead among other algorithms.}
    \end{figure*}
    
    \begin{figure*}[htb]
      \begin{subfigure}{0.32\textwidth}
        \includegraphics[width=\linewidth]{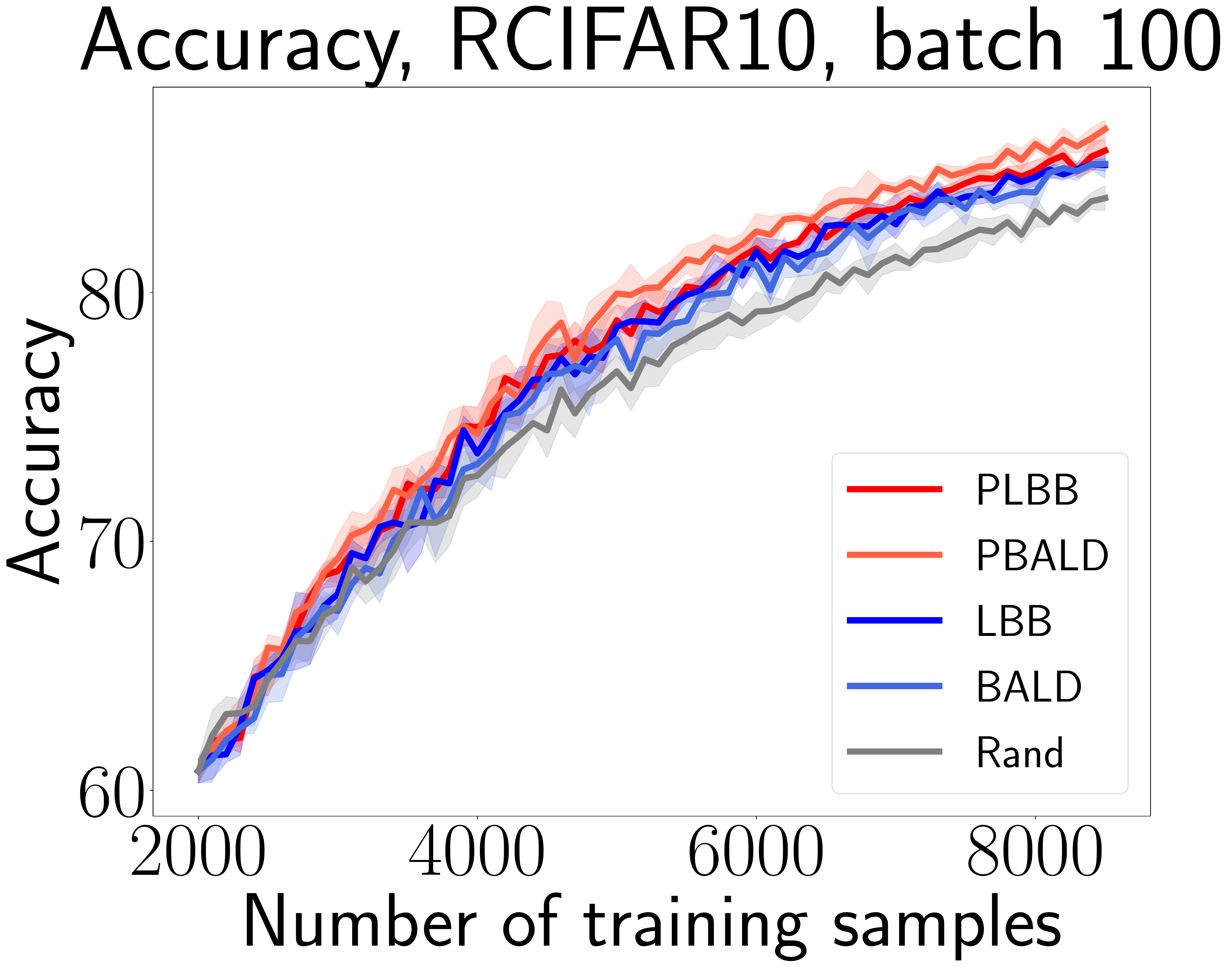}
        \caption{} \label{fig:ens_rcifar10_batch100}
      \end{subfigure}%
      \hspace*{\fill}
      \begin{subfigure}{0.32\textwidth}
        \includegraphics[width=\linewidth]{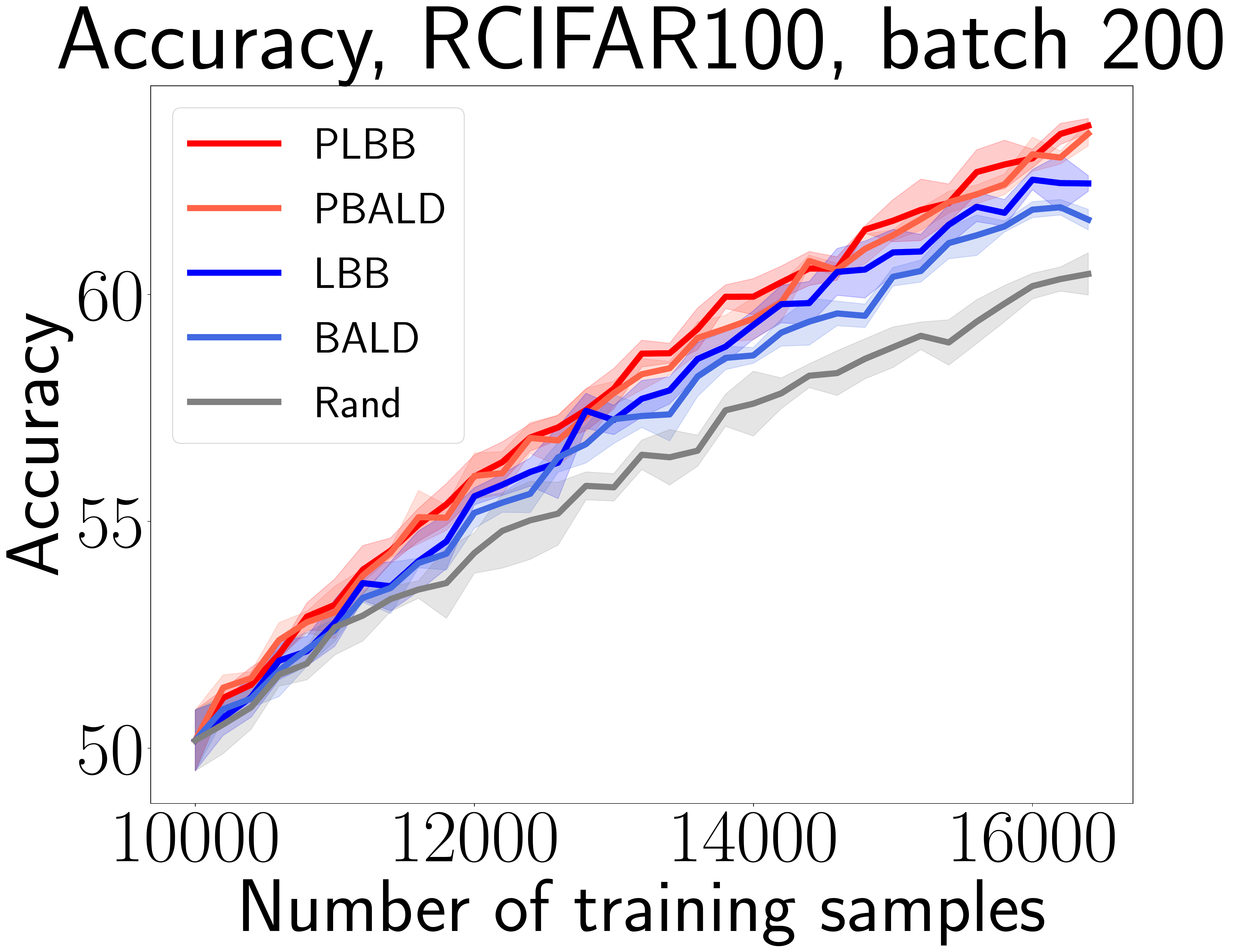}
        \caption{} \label{fig:ens_rcifar100_batch200}
      \end{subfigure}%
      \hspace*{\fill}
      \begin{subfigure}{0.32\textwidth}
      \includegraphics[width=\linewidth]{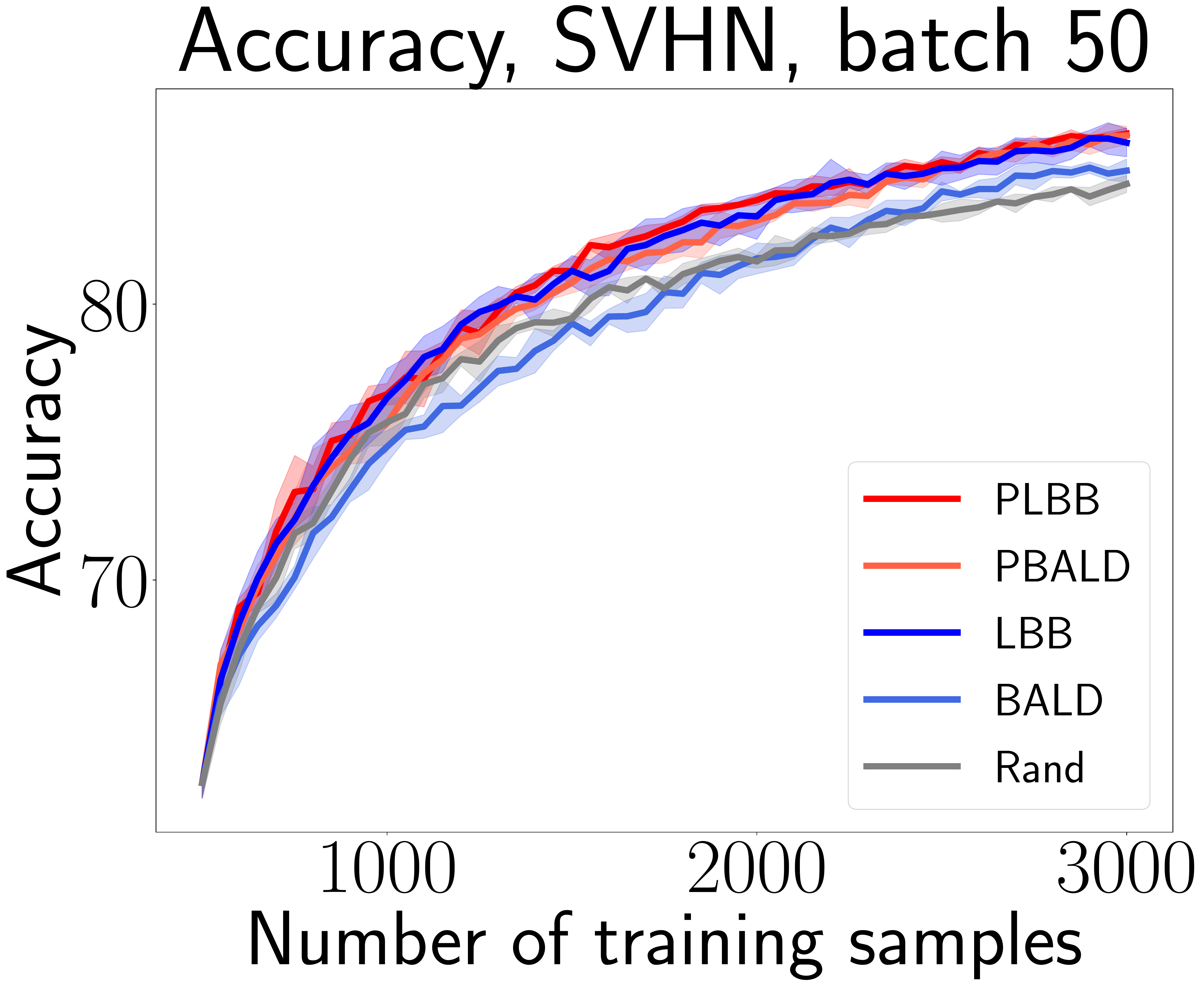}
        \caption{} \label{fig:ens_svhn_batch50}
      \end{subfigure}
    
      \caption{Performance comparison of AL algorithms on deep ensembles. Datasets: (a)~RCIFAR-10. (b)~RCIFAR-100. (c)~SVHN. LBB outperforms BALD, and also shows results close to the leading methods based on power distribution, namely PLBB and PBALD.}
    \end{figure*}
    
    

\subsubsection{Results}    
    The results of the experiment on the MNIST dataset are shown in Figure~\ref{fig:ens_mnist_batch10}.
    The Large BatchBALD algorithm is slightly better than the BALD algorithm, and as a BatchBALD approximation it is quite close to the original.
    As for the LBB and BALD extensions, namely PLBB and PBALD, they dominate among other algorithms, even outperforming the BatchBALD algorithm in both the ensemble and MC-dropout cases.

    In the RMNIST experiments, see Figure~\ref{fig:ens_rmnist_batch10}, BALD takes many similar images that do not introduce significant diversity into the training dataset, which is reflected in a loss of quality compared to other algorithms.
    Large BatchBALD, on the other hand, as an approximation of BatchBALD performs much better than BALD, taking into account batch interconnections.
    At first, it is slightly inferior to the random baseline, but then outperforms it, starting with a few hundred elements.
    In turn, PLBB and PBALD, which combine informativity proportional to the LBB and BALD criterion scores, respectively, and the diversity obtained by sampling from the power distribution, are quite close to each other in their performance and outperform all other algorithms.
    Note that on a dataset with a large pool, like RMNIST, and on experiments with large batches, BatchBALD becomes computationally infeasible.

    Regarding the results on the FMNIST dataset, the Large BatchBALD algorithm performs better than BALD. 
    The results of its PowerLBB and PowerBALD extensions are the best among other algorithms, with PLBB significantly outperforming PBALD, see Figure~\ref{fig:ens_fmnist_batch10}.
    This may be due to the fact that PLBB has the additional data diversity contained in the LBB algorithm design, while PBALD has data diversity only due to sampling-driven randomness.
    Both mentioned algorithms are better than BatchBALD, which in turn, together with BALD, has comparable performance and worse results among competitors.
    There are also results based on MC-dropout and results with larger batch sizes on the same datasets, see Supplementary Material, Sections~\ref{sec:mc_dropout} and~\ref{sec:bigger_batch_results}, respectively.
    Also, results for the Extended MNIST (EMNIST)~\cite{cohen2017emnist} and Kuzushiji-MNIST (KMNIST)~\cite{clanuwat2018deep} datasets in terms of Dolan-More curves~\cite{dolan_more_paper} can be found in Supplementary Material, Section~\ref{sec:dolan_more}.

\subsection{SVHN, RCIFAR-10, RCIFAR-100}
    CIFAR-10 and CIFAR-100~\cite{krizhevsky2009learning} are datasets of color images, each containing $60,000$ images consisting of $10$ and $100$ classes, respectively.
    Repeated extensions of CIFAR datasets involve repeating each of the images in the dataset multiple times with a Gaussian noise applied, namely $4$ times in our case, which increases the total size of each dataset proportionally.
    SVHN~\cite{netzer2011reading} is a Street View House Numbers dataset containing $70,000$ images.
    
    As a model, we used ResNet-18~\cite{he2016deep} architecture with an SGD optimizer with momentum $= 0.9$, weight decay $=0.0005$, learning rate $= 0.05$.
    As a learning rate scheduler, we used MultiStepLR with gamma $=0.1$, and milestones $= 25, 40$.
    The network was trained over $50$ epochs, and the version that showed the best quality on the validation set (the validation set consists of 5K examples) was used.
    
    The results for the SVHN dataset are presented in Figure~\ref{fig:ens_svhn_batch50} for a batch size $50$. LBB algorithm shows superiority over the BALD algorithm and the random baseline, while the BALD algorithm is inferior to the random baseline up to 2K examples.
    Additional randomization significantly improves the performance of the BALD algorithm, as PBALD shows.
    At the same time, LBB is as good as, and in some places slightly better than, the PBALD algorithm without additional randomization.
    The leader among all algorithms is the randomized version of LBB, the PLBB algorithm.
    
    The Repeated CIFAR-100 (RCIFAR-100) dataset is very challenging for the task of active learning due to the large number of classes and image repetitions, which increases the initial volume by $4$ times.
    Moreover, such a dataset requires taking samples in large batches to get good performance in a reasonable amount of time.
    Note that while the results based on RCIFAR-10 (see Figure~\ref{fig:ens_rcifar10_batch100}) show only slight superiority of the LBB algorithm over the BALD algorithm, on a more complex dataset like RCIFAR-100 the differences are much more clear, see Figure~\ref{fig:ens_rcifar100_batch200}.
    It shows that Large BatchBALD is more successful in quality than BALD and random baseline.
    As for the randomized versions of LBB and BALD, namely PLBB and PBALD, they improve the results of the original, with PLBB dominating in quality among the other algorithms on this dataset.
    For experimental results on the mentioned datasets for larger batch sizes, see Supplementary Material, Section~\ref{sec:bigger_batch_results}.

\subsection{Text-domain experiments}
\label{sec:text_experiments}
    For a more complete analysis, we also tested the proposed method on text data.
    We considered AG News~\cite{zhang2015character}, 
    which is a large dataset including 120K sentences from news articles attributed to $4$ classes.
    In this series of experiments, we use the DistilBERT~\cite{sanh2019distilbert} model as a one capable of obtaining acceptable quality with minimal runtime.
    Here we use MC-dropout to estimate uncertainty.
    
\begin{figure}[t!]
        \centering
        \includegraphics[width=0.75\linewidth]{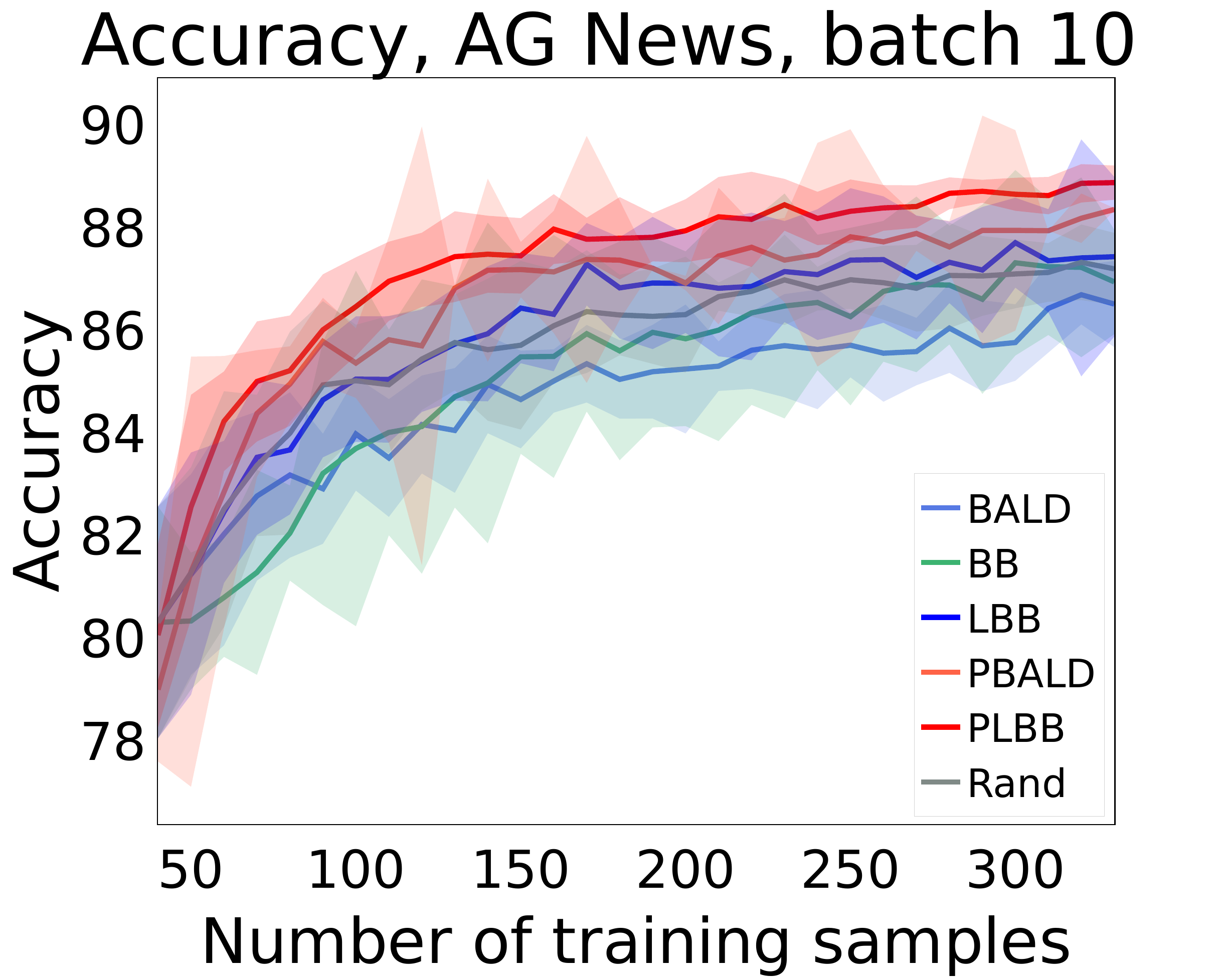}
        \caption{Performance comparison of AL algorithms on AG News dataset. PLBB is a top-performer, as well as PBALD and LBB clearly outperform other competitors.} \label{fig:ag}
    \end{figure}

    The results on AG News dataset show that the proposed PLBB algorithm shows the best quality among the competitors, see Figure~\ref{fig:ag}. 
    Moreover, even the original LBB performs better than the naive baselines, as well as BALD and BatchBALD.
    Note that this behavior is noticeable even on a small $10$-sample batch size.
    There are also results on a bigger batch size and with more competitors,
    see Supplementary Material, Section~\ref{sec:appendix_texts}.

\subsection{Algorithm runtime comparison}
    We present numerical execution times for the considered algorithms, namely BALD, PowerBALD, BatchBALD, Large BatchBALD, and Power Large BatchBALD, see Table~\ref{tab:ens_mnist_time_comparison} 
    which is based on MNIST dataset.
    Execution results are obtained with deep ensembles constructed using a small convolutional neural network.
    The initial pool consists of $20$ images, and the unlabeled pool contains $49,880$ images.
    Note that these results also support the claim that LBB, being an approximation of BatchBALD, is multiple times faster.
    Moreover, this difference becomes even more noticeable as the batch size increases, which can give a gain of tens of times.
    Thus, when working with batches of hundreds of items, it is evident that the calculation of the Large BatchBALD acquisition function is more feasible than that of the BatchBALD method.
    
    \begin{table*}[ht]
        \centering
        \begin{tabular}{|c|c|c|c|c|c|c|}
          \hline
          Batch size & BALD & PowerBALD & BatchBALD & Large BB & Power LBB \\ \hline
          10 & $5.04 \pm 0.51$ & $5.29 \pm 0.49$ & $268.9 \pm 10.37$ & $18.18 \pm 1.22$ & $20.13 \pm 2.61$ \\ \hline
          20 & $5.13 \pm 0.43$ & $5.93 \pm 1.43$ & $838.85 \pm 98.22$ & $20.06 \pm 5.31$ & $18.56 \pm 2.27$ \\ \hline
        \end{tabular}
        \caption{Algorithm runtime comparison in seconds. MNIST dataset, unlabeled pool of $49,980$ images, uncertainty estimation with deep ensembles. The proposed LBB-based approximation allows a significant reduction in computational cost compared to BatchBALD, even on batches of tens of elements.}
        \label{tab:ens_mnist_time_comparison}
    \end{table*}


\section{Related work}
\label{sec:related_work}
    Uncertainty estimates in deep learning are often associated either with MC-dropout~\cite{gal2016mcdropout} or deep ensembles~\cite{lakshminarayanan2017ensembles}.
    In the case of MC-dropout, one samples the dropout mask on inference to get an ensemble of models differently parameterized.
    In the case of deep ensembles, one trains a single model with different weight initialization.
    Speaking of classification, in both scenarios, the final prediction is the average of the softmax vectors from all the models in the ensemble.
    In the work~\cite{beluch2018ensembles} authors demonstrate the superior performance of ensembles over MC-dropout in image classification tasks.
    Nevertheless, we tested both of the approaches for dealing with uncertainty.
    
    One of the most applicable baseline algorithms for active learning is Bayesian Active Learning by Disagreement (BALD)~\cite{houlsby2011bald}.
    Its acquisition function is computed as mutual information between model output and its latent parameters.
    That is, BALD tries to find those examples in which the different models disagree, while each of the models is confident in its prediction.
    This approach is quite efficient when taking one example at each step for training.
    
    In practice, with a large pool size, it is unprofitable to take a single example or even small batches, so it is important to be able to take batches of informative examples at each step of the AL loop. 
    In practice, the top-$k$ approach is most often used to take more than one example in the AL loop, where each step takes $k$ examples with the highest values of the acquisition function.
    This approach has a serious drawback, namely, it does not take into account pairwise interaction of the data samples in batches that leads to acquiring a batch of similar examples and resulting performance degradation.

    One possible solution is a batch modification of the BALD algorithm called BatchBALD~\cite{gal2019batchbald}.
    The idea is to treat the mutual information in a batch manner as between a joint random variable (i.~e., set of model outputs) and model parameters.
    In this scenario, points are added one at a time in a greedy manner, and the total mutual information is recursively recalculated.
    The diversity of acquired samples comes from accounting for the interactions in the batch between outputs.
    Nevertheless, BatchBALD has a significant drawback, namely, it takes a lot of working time~\cite{kirsch2021stoch_batch}.
    In the original BatchBALD work, the standard choice is to take a batch of 10 elements, since complexity grows exponentially with batch size due to the joint entropy calculation.
    In practice, it is calculated directly for the first 5 samples in the batch, and the rest are sampled using MC-dropout.
    
    Another way for diversification is presented in the already mentioned article~\cite{kirsch2021stoch_batch}.
    Its authors get a diversity of chosen examples at the expense of the stochasticity of the acquisition function.
    Their idea is to get scores from some algorithm, such as BALD, and then translate those scores into a distribution.
    Sampling from such a distribution, we get an acquisition batch.
    Thus, if we take all the scores obtained with BALD, raise to some power, normalize, and then sample, we obtain the PowerBALD algorithm, which is one of the comparative baselines in this paper.
    The basic idea is that a randomized sampling strategy is better than a greedy one, requires the same amount of time, and partially overcomes the data redundancy bottleneck.
    
    To achieve the best quality of AL algorithm, it is often useful to focus only on uncertainty that is directly related to the quality of the algorithm.
    Thus, the authors of MOCU-based algorithms~\cite{zhao2021smocu} propose to minimize only the classification error uncertainty as an acquisition function, taking into account the posterior, rather than the overall uncertainty, in contrast to BALD.
    As a result, the authors do not increase the probability of an already guessed classes, but extract controversial samples on the classification borders, taking into account the posterior.
    A similar idea was considered in the paper~\cite{panov2015adoe} where the authors proposed an acquisition function in a one-step look-ahead manner for regression on Gaussian processes~\cite{rasmussen2006gaussian}.
    The idea is to choose a new point to add so that the average variance over the space is reduced.
    
    Another important issue in active learning, affecting the performance of the model, is the diversity of acquired data samples.
    Combining a Bayesian network and a Gaussian process with a known covariance function, the authors in~\cite{tsymbalov2019alnngp} propose to obtain diverse data samples from the acquisition function as the maximum variance of the Gaussian process.
    Also, since the variance of the Gaussian process does not depend on the output of the network, each time sampling a new point changes the total variance, which eliminates the need to retrain the network at each step.
    Another design of the acquisition function, which takes into account the diversity of samples, is based on the geometric properties of the data~\cite{sener2018active}.
    The idea is to add images with the greatest distance from the training set to find data that is still poorly represented by the training set.
    
    Another natural but computationally expensive way to introduce diversity of AL samples into a training set is to use clustering.
    The authors in the paper~\cite{citovsky2021batch} demonstrate an efficient data sampling with huge batch sizes by selecting samples from hierarchically clustered data in an ascending volume manner.
    Another current state-of-the-art work~\cite{ash2019badge} uses k-means++ to achieve diverse acquired samples, along with an acquisition function built on the value of loss gradients relative to model parameters as the value of potential model change.
    In the work~\cite{wan2021nearest} the authors suggest using KNN classifier as the output layer of the network instead of softmax, due to better generalization ability to the unknown space.


\section{Conclusions}
\label{sec:conclusions}
    To summarize, in this work we propose a new active learning algorithm \textit{Large BatchBALD} that performs an approximation of the BatchBALD method, using the BALD acquisition function and pairwise mutual information of model output components.
    The proposed algorithm is as efficient in avoiding taking similar objects in one batch as the original method, while it computes the acquisition function several times faster, especially in the case of large batches. 
    Thus, this active learning algorithm balances the uncertainty and diversity of the acquired samples and significantly reduces the acquisition time compared to the original BatchBALD.
    The resulting method is shown to be efficient for batch AL in application to modern image and text datasets.
    The code to reproduce the experiments is available online at~\url{https://github.com/stat-ml/large_batch_bald}.\\

    \noindent \textbf{Acknowledgments.} The research was supported by the Russian Science Foundation grant 20-71-10135.
    The authors acknowledge the use of computational resources of the Skoltech supercomputer Zhores~\cite{zacharov2019zhores} for obtaining the results presented in this paper.

\bibliographystyle{plain}
\bibliography{lit}

\clearpage

\newpage

\appendix

\section{Supplementary Material}
\label{sec:appendix_short}

\subsection{Main property proof}
\label{sec:proof}
    Let us prove the following statement:
    \begin{equation}
        \sum_{i=1}^b I(y_i, \theta) - I(y_{1:b}; \theta) = C(y_{1:b}).
    \end{equation}
    
    Mutual information (aka BALD acquisition function) can be written by definition:
    \begin{equation}
        \begin{aligned}
            I(y_i; \theta) &= \int_{\theta} \int_{y_i} p(y_i; \theta) \log \dfrac{p(y_i; \theta)}{p(y_i) p(\theta)} \mathrm{d} y_i \mathrm{d} \theta \\
            &= \int_{\theta} \int_{y_i} p(y_i; \theta) \log \dfrac{p(y_i \mid \theta)}{p(y_i)} \mathrm{d} y_i \mathrm{d} \theta. 
        \end{aligned}
    \end{equation}
    
    Mutual information of a joint random variable (aka BatchBALD acquisition function) can be expressed as:
    \begin{equation*}
        \begin{aligned}
            I(y_{1:b}; \theta) &= \int_{\theta} \int_{y_{1:b}} p(y_{1:b}; \theta) \log \dfrac{p(y_{1:b}; \theta)}{p(y_{1:b}) p(\theta)} \mathrm{d} y_{1:b} \mathrm{d} \theta \\
            &= \int_{\theta} \int_{y_{1:b}} p(y_{1:b} \mid \theta) p(\theta) \log \dfrac{p(y_{1:b} \mid \theta)}{p(y_{1:b})} \mathrm{d} y_{1:b} \mathrm{d} \theta  \\
            &= \int_{\theta} \int_{y_{1:b}} p(y_{1:b} \mid \theta) p(\theta) \log p(y_{1:b} \mid \theta) \mathrm{d} y_{1:b} \mathrm{d} \theta \\
            &- \int_{\theta} \int_{y_{1:b}} p(y_{1:b} \mid \theta) p(\theta) \log p(y_{1:b}) \mathrm{d} y_{1:b} \mathrm{d} \theta  \\
            &= \int_{\theta} \int_{y_{1:b}} p(y_{1:b} \mid \theta) p(\theta) \log p(y_{1:b} \mid \theta) \mathrm{d} y_{1:b} \mathrm{d} \theta \\
            &- \int_{y_{1:b}} p(y_{1:b}) \log p(y_{1:b}) \mathrm{d} y_{1:b}.
        \end{aligned}
    \end{equation*}
    
    Mutual information of $b$ random variables one can write as:
    \begin{equation*}
        \begin{aligned}
            &C(y_{1:b}) = \int \limits_{y_1:b} p(y_{1:b}) \log \dfrac{p(y_{1:b})}{\prod \limits_{i=1}^{b} p(y_i)} \mathrm{d} y_{1:b} \\
            &= \int \limits_{y_{1:b}} p(y_{1:b}) \log p(y_{1:b}) \mathrm{d} y_{1:b} - \int \limits_{y_{1:b}} p(y_{1:b}) \log \prod \limits_{i=1}^{b} p(y_i) \mathrm{d} y_{1:b} \\
            &= \int \limits_{y_{1:b}} p(y_{1:b}) \log p(y_{1:b}) \mathrm{d} y_{1:b} \\
            &- \sum \limits_{i=1}^{b} \int \limits_{y_{1:b}} p(y_{1:b} \mid \theta) p(\theta) \log p(y_i) \mathrm{d} y_{1:b} \\
            &= \int \limits_{y_{1:b}} p(y_{1:b}) \log p(y_{1:b}) \mathrm{d} y_{1:b} - \sum \limits_{i=1}^{b} \int \limits_{y_{i}} p(y_{i}) \log p(y_i) \mathrm{d} y_{i}. \\
        \end{aligned}
    \end{equation*}
    Using the equations above, we can write the following:
    \begin{equation}
        \begin{aligned}
            & \sum_{i=1}^b I(y_i; \theta) - I(y_{1:b}; \theta) - C(y_{1:b}) \\
        \end{aligned}
    \end{equation}
    \begin{equation*}
        \begin{aligned}
            & \textrm{(by definition)} \\
            =& \sum_{i=1}^b \int_{\theta} \int_{y_i} p(y_{i}; \theta) \log \dfrac{p(y_i; \theta)}{p(y_i) p(\theta)} \mathrm{d} y_i \\
            -& \int_{\theta} \int_{y_{1:b}} p(y_{1:b}; \theta) \log \dfrac{p(y_{1:b}; \theta)}{p(y_{1:b}) p(\theta)} \mathrm{d} y_{1:b} \\
            -& \int_{y_{1:b}} p(y_{1:b}) \log \dfrac{p(y_{1:b})}{\prod \limits_{i=1}^{b} p(y_i)} \mathrm{d} y_{1:b}
        \end{aligned}
    \end{equation*}
    \begin{equation*}
        \begin{aligned}
            & \textrm{(reduce the numerator and denominator)} \\
            =& \sum_{i=1}^b \int_{\theta} \int_{y_i} p(y_{i}; \theta) \log \dfrac{p(y_i \mid \theta)}{p(y_i)} \mathrm{d} y_i \\
            -& \int_{\theta} \int_{y_{1:b}} p(y_{1:b}; \theta) \log \dfrac{p(y_{1:b} \mid \theta)}{p(y_{1:b})} \mathrm{d} y_{1:b} \\
            -& \int_{y_{1:b}} p(y_{1:b}) \log \dfrac{p(y_{1:b})}{\prod \limits_{i=1}^{b} p(y_i)} \mathrm{d} y_{1:b}
        \end{aligned}
    \end{equation*}
    \begin{equation*}
        \begin{aligned}
            & \textrm{(factorization of joint probability due to} \\ 
            & \textrm{independence of $y_i$ conditioned on $\theta$)} \\
            =& \sum_{i=1}^b \int_{\theta} \int_{y_i} p(y_{i}; \theta) \log \dfrac{ p(y_{i} \mid \theta)}{ p(y_i)} \mathrm{d} y_i \\
            -& \int_{\theta} \int_{y_{1:b}} p(y_{1:b}; \theta) \log \dfrac{\prod \limits_{i=1}^b p(y_i \mid \theta)}{p(y_{1:b})} \mathrm{d} y_{1:b} \\
            -& \int_{y_{1:b}} p(y_{1:b}) \log \dfrac{p(y_{1:b})}{\prod \limits_{i=1}^{b} p(y_i)} \mathrm{d} y_{1:b}
        \end{aligned}
    \end{equation*}
    \begin{equation*}
        \begin{aligned}
            & \textrm{(log of product is equal to sum of logs} \\
            & \textrm{and log fraction is expressed as a diff. of logs)} \\
            =& \sum_{i=1}^b \int_{\theta} \int_{y_i} \bigl( p(y_{i} \mid \theta) p(\theta) \log p(y_{i} \mid \theta) \bigr. \\
            -& \bigl. p(y_{i} \mid \theta) p(\theta) \log p(y_i) \bigr) \mathrm{d} y_i \mathrm{d} \theta \\
            -& \sum_{i=1}^b \int_{\theta} \int_{y_{1:b}} \bigl( p(y_{1:b} \mid \theta) p(\theta) \log p(y_i \mid \theta) \bigr. \\
            -& \bigl. p(y_{1:b} \mid \theta) p(\theta) \log p(y_{1:b}) \bigr) \mathrm{d} y_{1:b} \mathrm{d} \theta \\
            -& \int_{y_{1:b}} p(y_{1:b}) \log \dfrac{p(y_{1:b})}{\prod \limits_{i=1}^{b} p(y_i)} \mathrm{d} y_{1:b}
        \end{aligned}
    \end{equation*}
    \begin{equation*}
        \begin{aligned}
            & \left(\textrm{integr. out: $ p(y_{1:b}; \theta) = \prod_{i=1}^b p(y_{i} \mid \theta) p(\theta)$}\right) \\
        \end{aligned}
    \end{equation*}
    \begin{equation*}
        \begin{aligned}
            =& \sum_{i=1}^b \int_{\theta} \int_{y_i} p(y_{i} \mid \theta) p(\theta) \log p(y_{i} \mid \theta) \mathrm{d} y_i \mathrm{d} \theta \\
            -& \sum_{i=1}^b \int_{y_i} p(y_{i}) \log p(y_i) \mathrm{d} y_i \\
            -& \sum_{i=1}^b \int_{\theta} \int_{y_{i}} p(y_{i} \mid \theta) p(\theta) \log p(y_i \mid \theta) \mathrm{d} y_{i} \mathrm{d} \theta \\
            +& \int_{y_{1:b}} p(y_{1:b}) \log p(y_{1:b}) \mathrm{d} y_{1:b}\\
            -& \int_{y_{1:b}} p(y_{1:b}) \log p(y_{1:b}) \mathrm{d} y_{1:b} + \sum_{i=1}^b \int_{y_i} p(y_{i}) \log p(y_i) \mathrm{d} y_i\\
            & = 0. \\
        \end{aligned}
    \end{equation*}
    which is an exact statement that was presented in the beginning.

\subsection{Pairwise mutual information calculation}
\label{sec:mut_info_calc}
    Here we give an explicit formula for calculating the pairwise mutual information $I(y_i; y_l)$:
    \begin{equation*}
        \begin{aligned}
            & I(y_i; y_l) = \sum_{\hat{y}_{i}, \hat{y}_{ l} = 1}^C \left[\frac{1}{k} \sum_{j=1}^{k} p\left(\hat{y}_{i} \mid \hat{\theta}_{j}
            \right) p\left(\hat{y}_{l} \mid \hat{\theta}_{j}
            \right)\right] \cdot \\
            & \quad \cdot \left[\log \left(\frac{1}{k} \sum_{j=1}^{k} p\left(\hat{y}_{i} \mid \hat{\theta}_{j}
            \right) p\left(\hat{y}_{l} \mid \hat{\theta}_{j}
            \right)\right)\right. \\
            & - \log \left. \left(\frac{1}{k} \sum_{j=1}^{k} p\left(\hat{y}_{i} \mid \hat{\theta}_{j}
            \right)\right) - \log \left(\frac{1}{k} \sum_{j=1}^{k} p\left(\hat{y}_{l} \mid \hat{\theta}_{j} \right)\right)\right],
        \end{aligned}
    \end{equation*}
    where $i \neq l$, and $k$ is either the number of forward passes in the case of MC-dropout, or the number of models in an ensemble in the case of deep ensembles.
    
    That is, it consists of two tensor multiplication operations with dimensionality $[n, k, C] \times [n, k, C] = [n, n, k, C]$ and $[n, C] \times [n, C] = [n, n, C]$, where $n$ is a processing batch size.

\subsection{Additional discussion of BALD-based algorithms computational complexity}
\label{sec:appendix_complexity}
  BatchBALD computational complexity can be described as follows.
  On each of $i = 1:b$ the acquisition steps, a new candidate $x_i$ with $p(y_i \mid \theta)$ from $|D_\text{pool}|$ is greedily selected to the already formed batch $p(y_{1:i-1} \mid \theta)$ of elements $x_{1:i-1}$.
  This batch is already calculated and stored, elements $x_{1:i-1}$ are fixed, so the task is to calculate joint entropy between a new added point $x_i$ 
  and an existed batch in a one by one manner.
  In exact (means based on given draws of $\theta$) joint entropy scenario, all possible combinations $y_{1:i-1}$ can be calculated exactly as $c^i$ meaning $c$ possible classes of each of $i$ elements in a batch.
  As for the approximated joint entropy scenario, if $c^i$ value is big (in BatchBALD paper it is assumed after $5$ acquired elements) then $p(y_{1:i-1})$ of $y_{1:i-1}$ is approximated using $m$ MC-samples.
  In both cases, joint probability $p(y_1, \ldots, y_i)$ is calculated by averaging over $p(y_1, \ldots, y_i \mid \theta)$ (i.~e., to find a probability density marginalizing over $\theta$) with $k$ MC-dropouts of $\theta$.
  So, batch is selected in linear time, although joint probability still requires a lot of computational resources both in exact and approximate setting.
    
  In a naive setting, complexity is $\mathcal{O}(c^b \cdot |D_\text{pool}|^b \cdot k)$ and can be described as follows.
  For every element of a batch of size $b$, from a data pool $|D_\text{pool}|$ with $c$ possible classes, a new data sample is searched as a maximum of difference between joint entropy and conditional joint entropy.
  The difference with efficient implementation is that in efficient option $x_{1:i-1}$ is fixed and varies only $x_i$ while in naive implementation all $x_{1:i}$ vary.
  $p(y_1, \ldots, y_b)$ is also calculated by averaging over $p(y_1, \ldots, y_b \mid \theta)$ with $k$ MC-samples.

\subsubsection{MC-dropout results}
\label{sec:mc_dropout}
\begin{figure*}
      \begin{subfigure}{0.32\textwidth}
        \includegraphics[width=\linewidth]{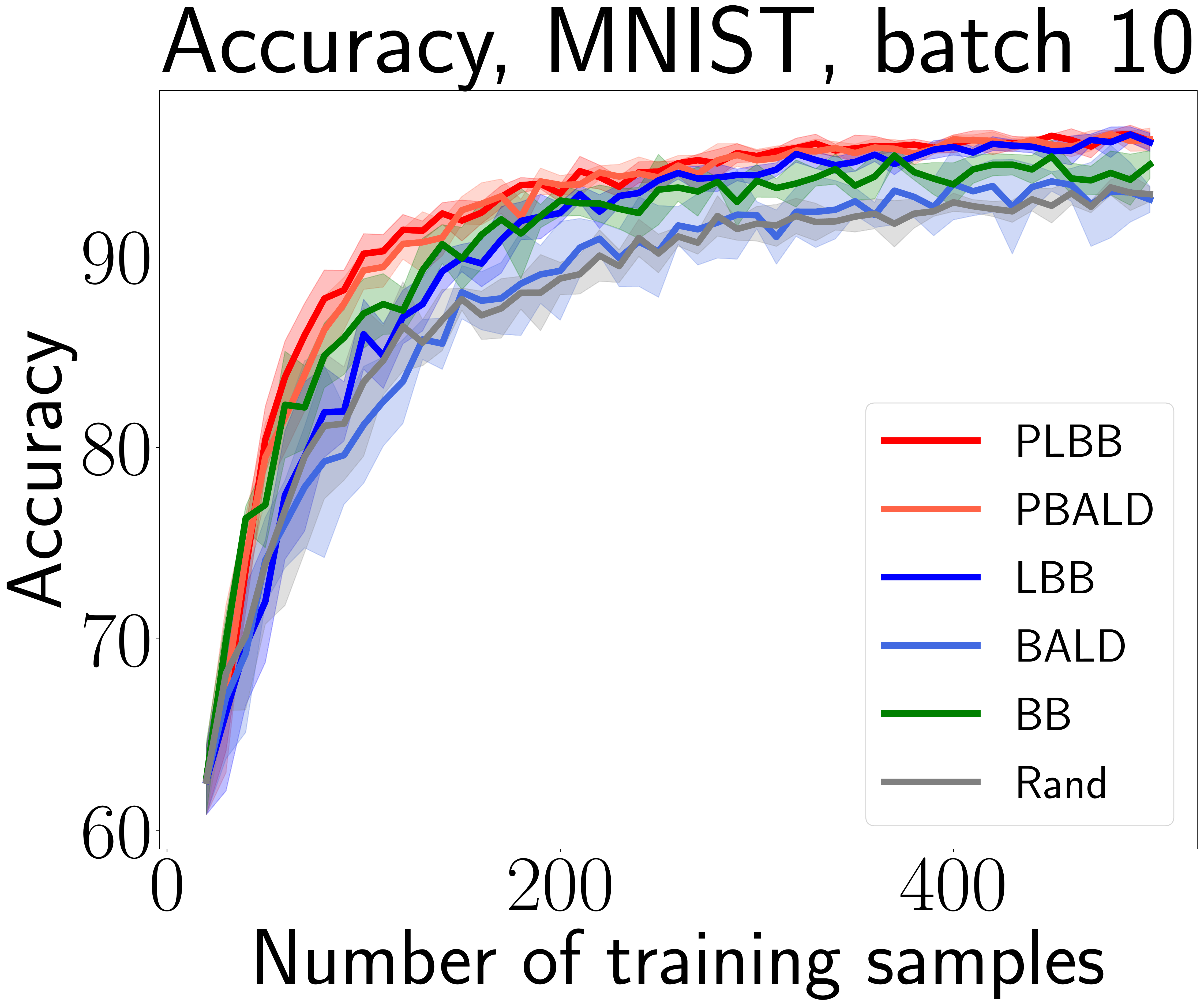}
        \caption{} \label{fig:mc_mnist_batch10}
      \end{subfigure}%
      \hspace*{\fill}
      \begin{subfigure}{0.32\textwidth}
          \includegraphics[width=\linewidth]{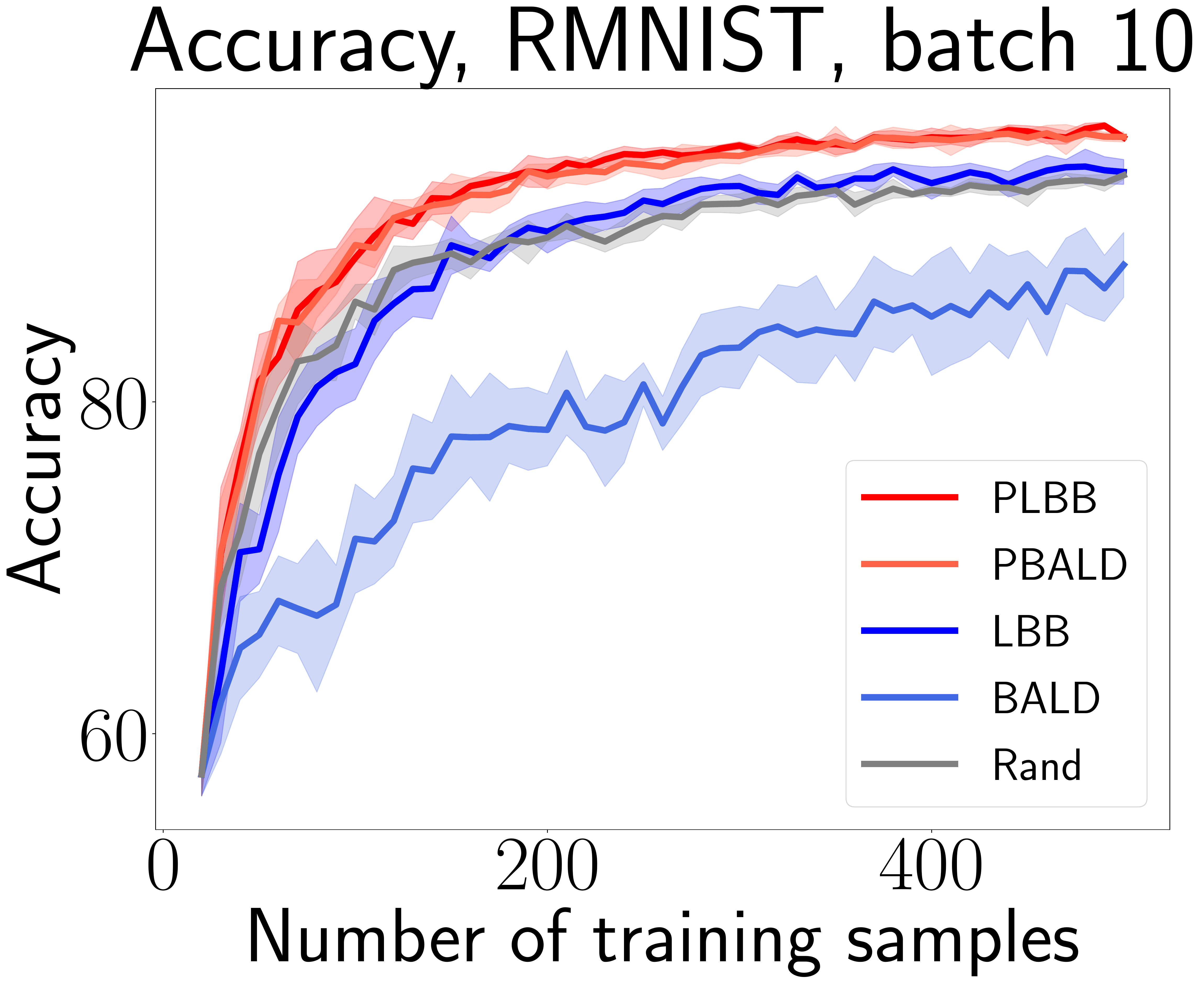}
        \caption{} \label{fig:mc_rmnist_batch10}
      \end{subfigure}
      \hspace*{\fill}
      \begin{subfigure}{0.32\textwidth}
        \includegraphics[width=\linewidth]{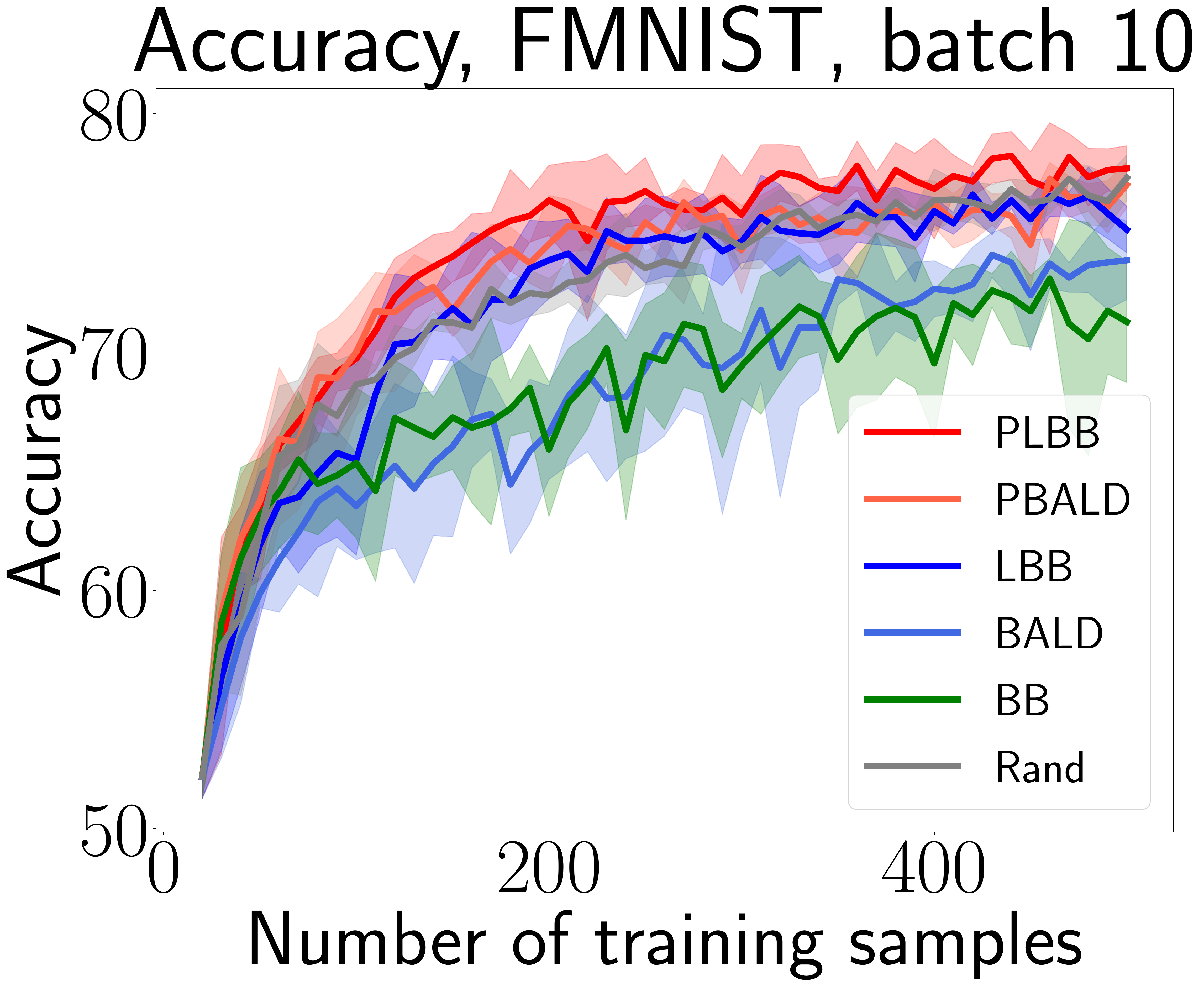}
        \caption{} \label{fig:mc_fmnist_batch10}
      \end{subfigure}
    
      \caption{Performance comparison of AL algorithms on MC-dropout. Datasets: (a)~MNIST. (b)~RMNIST. (c)~FMNIST. LBB is clearly superior to BALD, and by a larger margin than in the case of deep ensembles.}
    \end{figure*}

    Comparing the results obtained with MC-dropout and with deep ensembles, we see that the test accuracy of algorithms with MC-dropout is lower than with ensembles, as we mentioned earlier, see Figure~\ref{fig:mc_mnist_batch10}, Figure~\ref{fig:mc_rmnist_batch10}, Figure~\ref{fig:mc_fmnist_batch10}.
    At the same time, the margin between algorithms, for example, between LBB and BALD is more clear in the figures obtained with MC-dropout.
    Thus, the Large BatchBALD algorithm is significantly better than BALD.
    Also, in Figure~\ref{fig:mc_mnist_batch10} the LBB algorithm even outperforms BatchBALD in quality, starting from $200$ elements in the training set.
    The BALD algorithm, however, shows quality comparable to random selection of samples.
    Furthermore, in Figure~\ref{fig:mc_rmnist_batch10} and Figure~\ref{fig:mc_fmnist_batch10} the BALD performance is significantly worse than the random selection of samples.
    Also, in all three figures, power extensions show the best accuracy among all algorithms.
    
\subsection{Additional results for bigger batches}
\label{sec:bigger_batch_results}
\begin{figure*}
      \begin{subfigure}{0.32\textwidth}
        \includegraphics[width=\linewidth]{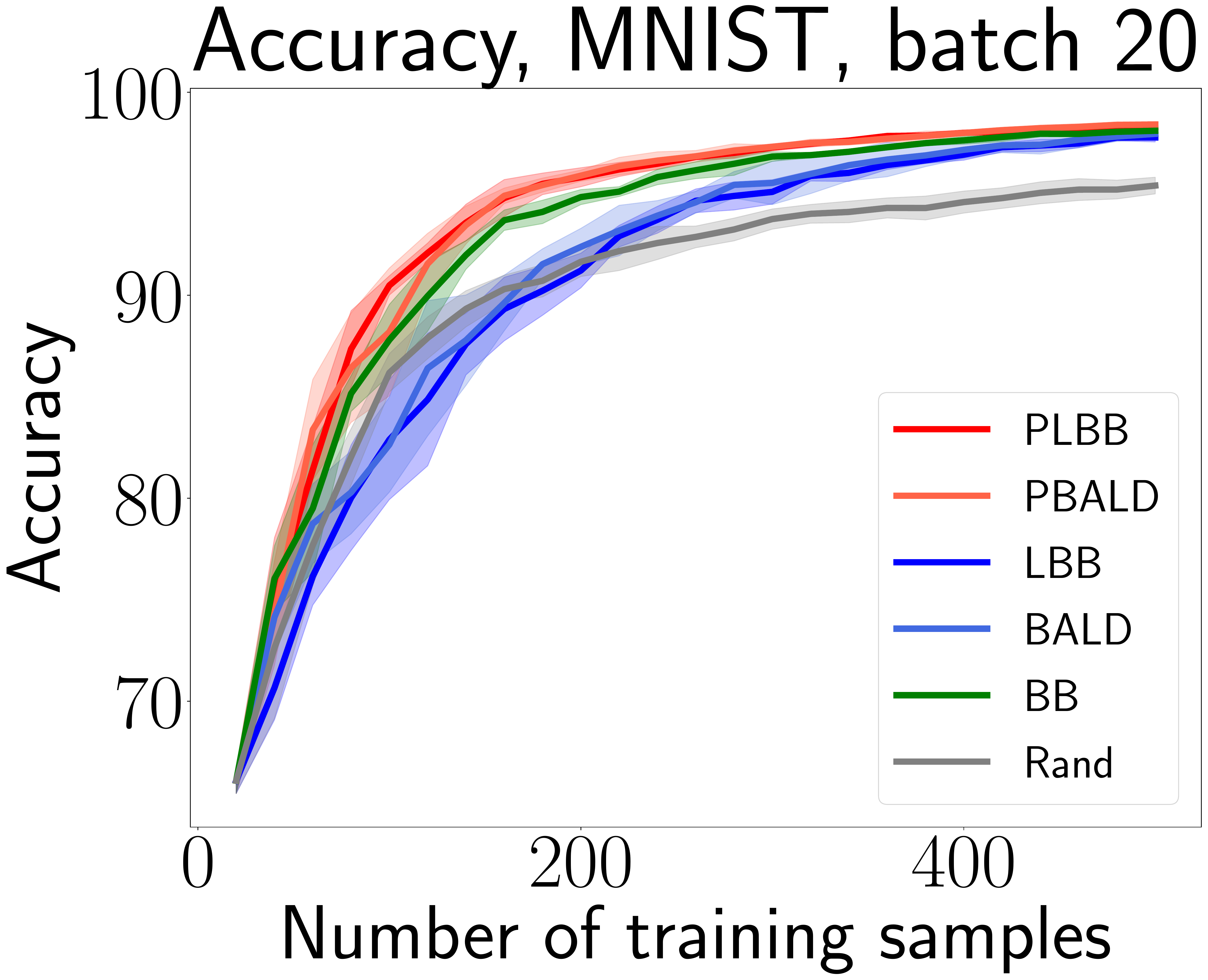}
        \caption{} \label{fig:ens_mnist_batch20}
      \end{subfigure}
      \hspace*{\fill}
      \begin{subfigure}{0.32\textwidth}
        \includegraphics[width=\linewidth]{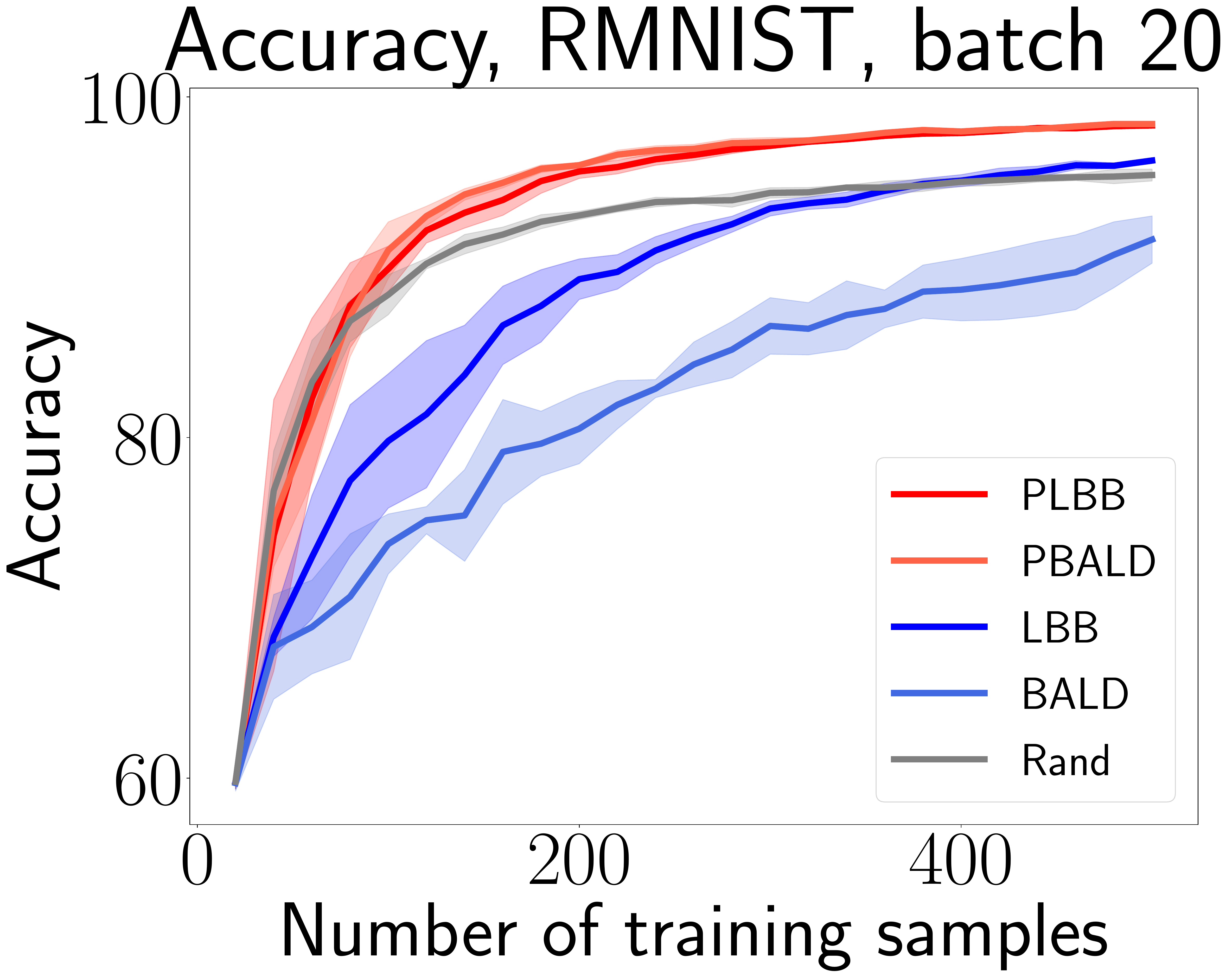}
        \caption{} \label{fig:ens_rmnist_batch20}
      \end{subfigure}%
      \hspace*{\fill}
      \begin{subfigure}{0.32\textwidth}
      \includegraphics[width=\linewidth]{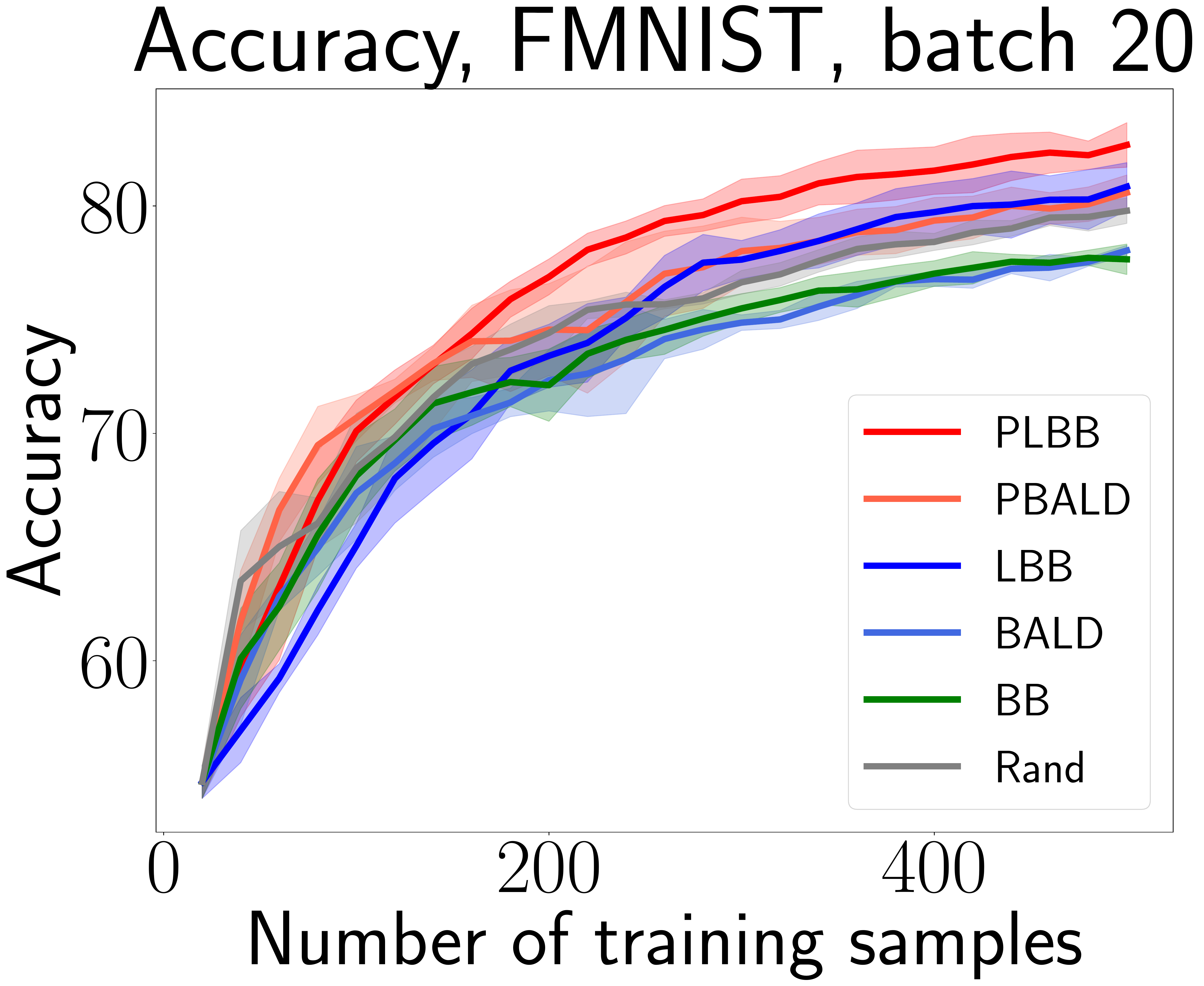}
        \caption{} \label{fig:ens_fmnist_batch20}
      \end{subfigure}
    
      \caption{Performance comparison of AL algorithms on deep ensembles. Datasets: (a)~MNIST. (b)~RMNIST. (c)~FMNIST.}
    \end{figure*}

\begin{figure*}
      \begin{subfigure}{0.32\textwidth}
        \includegraphics[width=\linewidth]{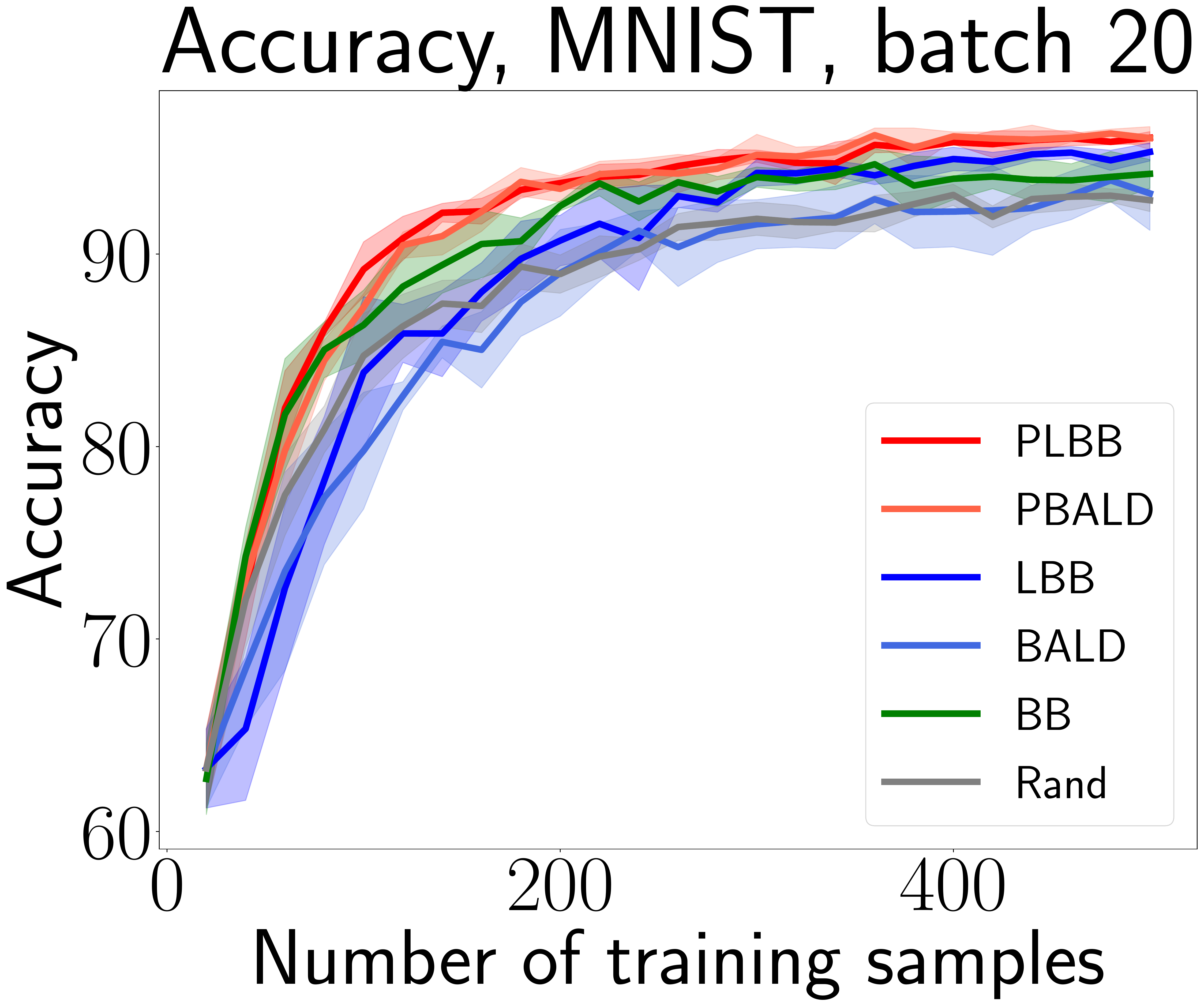}
        \caption{} \label{fig:mc_mnist_batch20}
      \end{subfigure}%
      \hspace*{\fill}
      \begin{subfigure}{0.32\textwidth}
          \includegraphics[width=\linewidth]{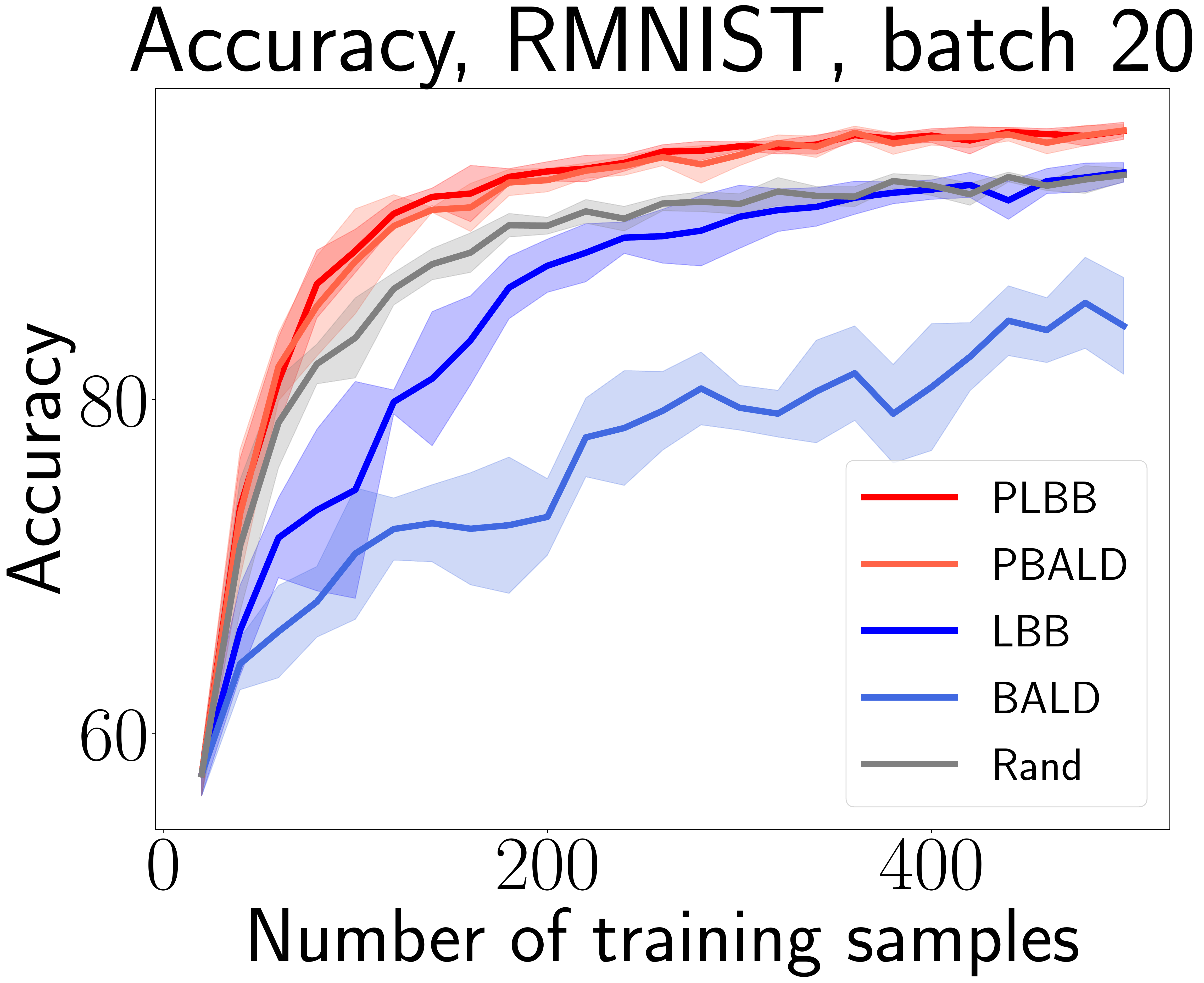}
        \caption{} \label{fig:mc_rmnist_batch20}
      \end{subfigure}
      \hspace*{\fill}
      \begin{subfigure}{0.32\textwidth}
        \includegraphics[width=\linewidth]{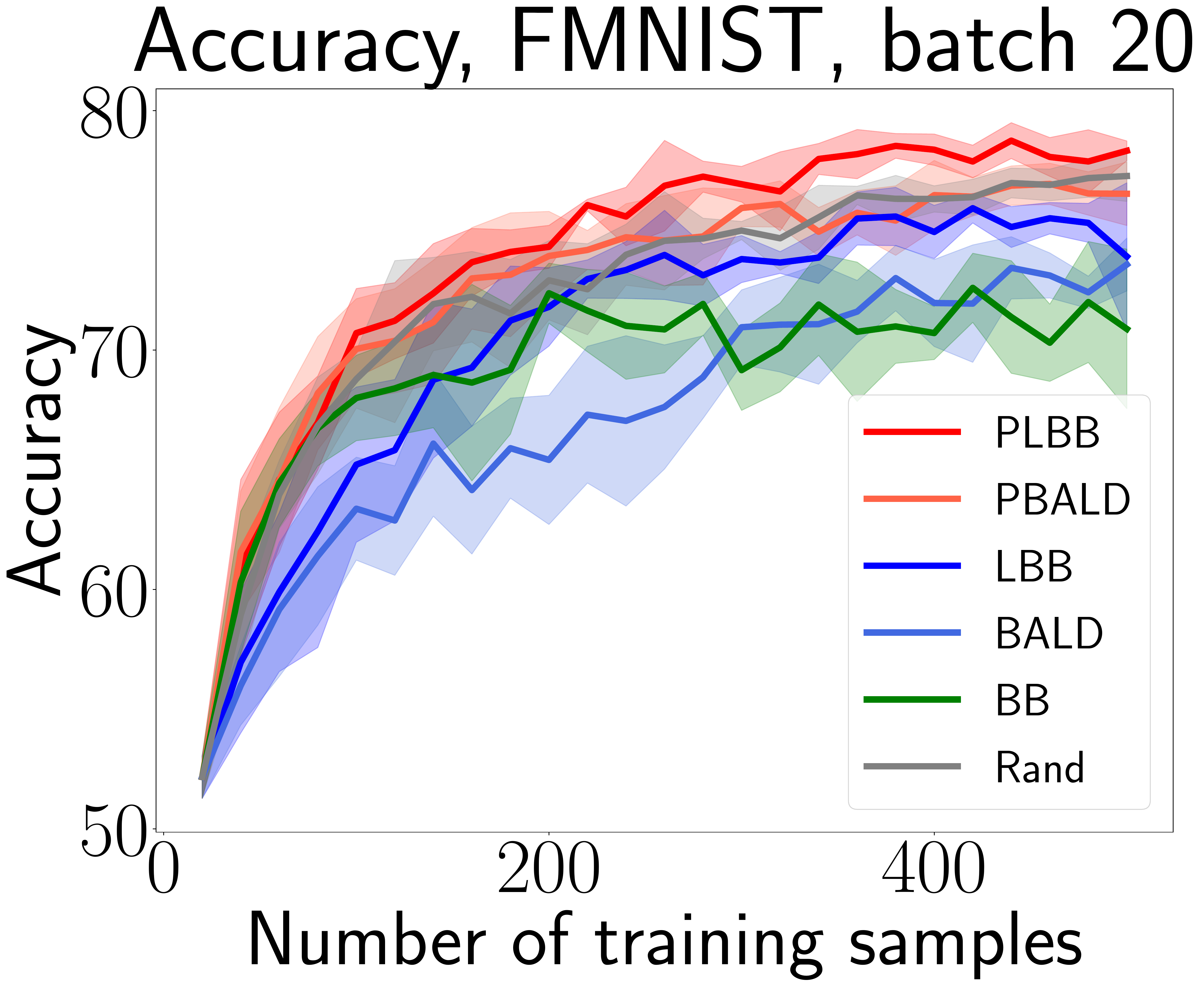}
        \caption{} \label{fig:mc_fmnist_batch20}
      \end{subfigure}
    
      \caption{Performance comparison of AL algorithms on MC-dropout. Datasets: (a)~MNIST. (b)~RMNIST. (c)~FMNIST. }
    \end{figure*}

    Experimental results on a larger acquisition batch for the MNIST dataset are shown in Figure~\ref{fig:ens_mnist_batch20}.
    Note that LBB and BALD have comparable accuracy here, as do PLBB and PBALD.
    Power extensions also dominate among other algorithms.
    The results for the RMNIST dataset with larger acquisition batch size are in Figure~\ref{fig:ens_rmnist_batch20}.
    Here we observe a similar picture for LBB and BALD, as well as for PLBB and PBALD, which in turn show the best quality among all algorithms.
    Similar conclusions can be drawn from the FMNIST dataset, with acquisition batch size equal to $20$, see Figure~\ref{fig:ens_fmnist_batch20}.
    For results on the MNIST, RMNIST, FMNIST datasets on the $20$ batch with MC-dropout, see Figure~\ref{fig:mc_mnist_batch20}, Figure~\ref{fig:mc_rmnist_batch20}, Figure~\ref{fig:mc_fmnist_batch20}, respectively.
    

     \begin{figure*}
      \begin{subfigure}{0.32\textwidth}
      \includegraphics[width=\linewidth]{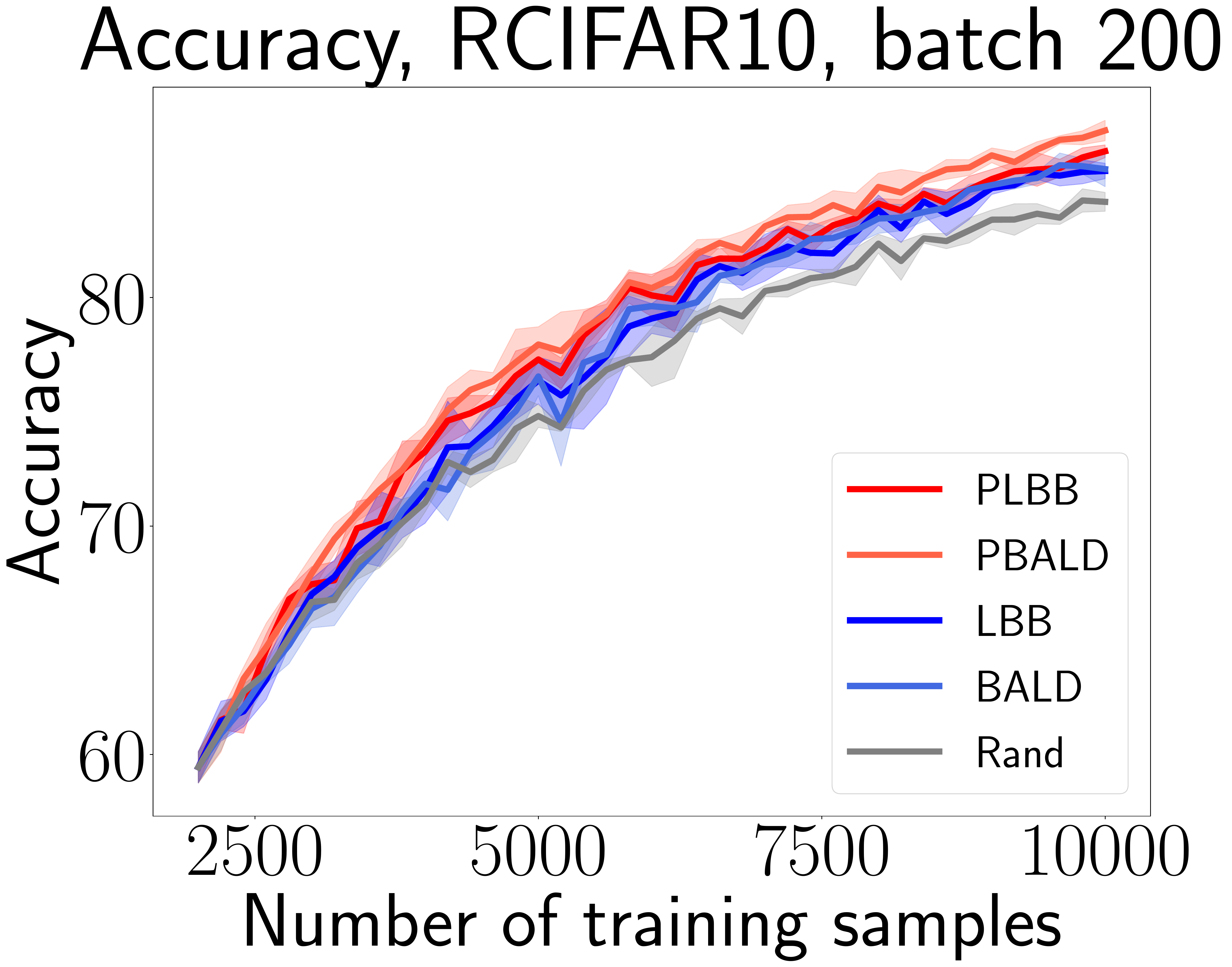}
        \caption{} \label{fig:ens_rcifar10_batch200}
      \end{subfigure}
      \hspace*{\fill}
      \begin{subfigure}{0.32\textwidth}
      \includegraphics[width=\linewidth]{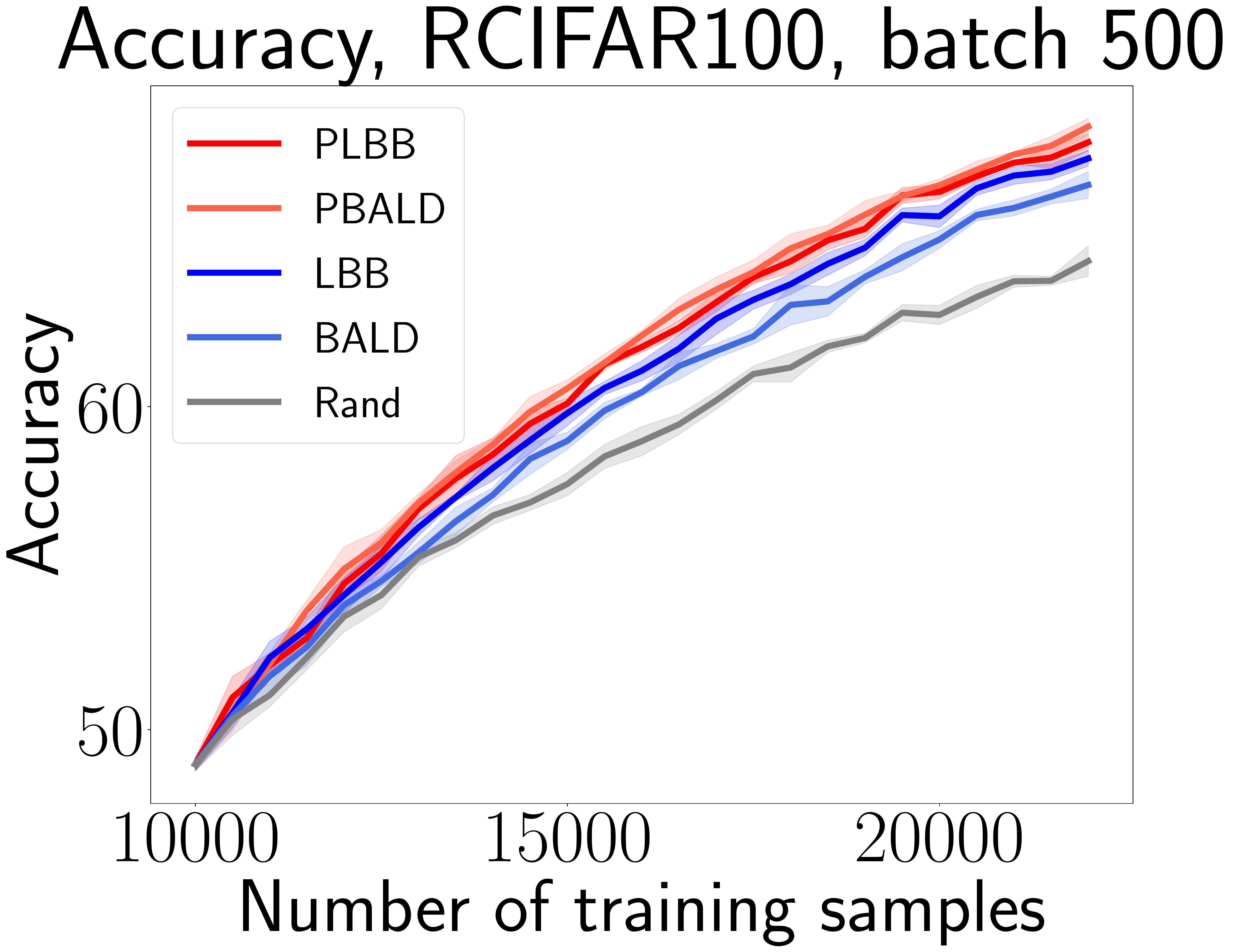}
        \caption{} \label{fig:ens_rcifar100_batch500}
      \end{subfigure}
      \hspace*{\fill}
      \begin{subfigure}{0.32\textwidth}
      \includegraphics[width=\linewidth]{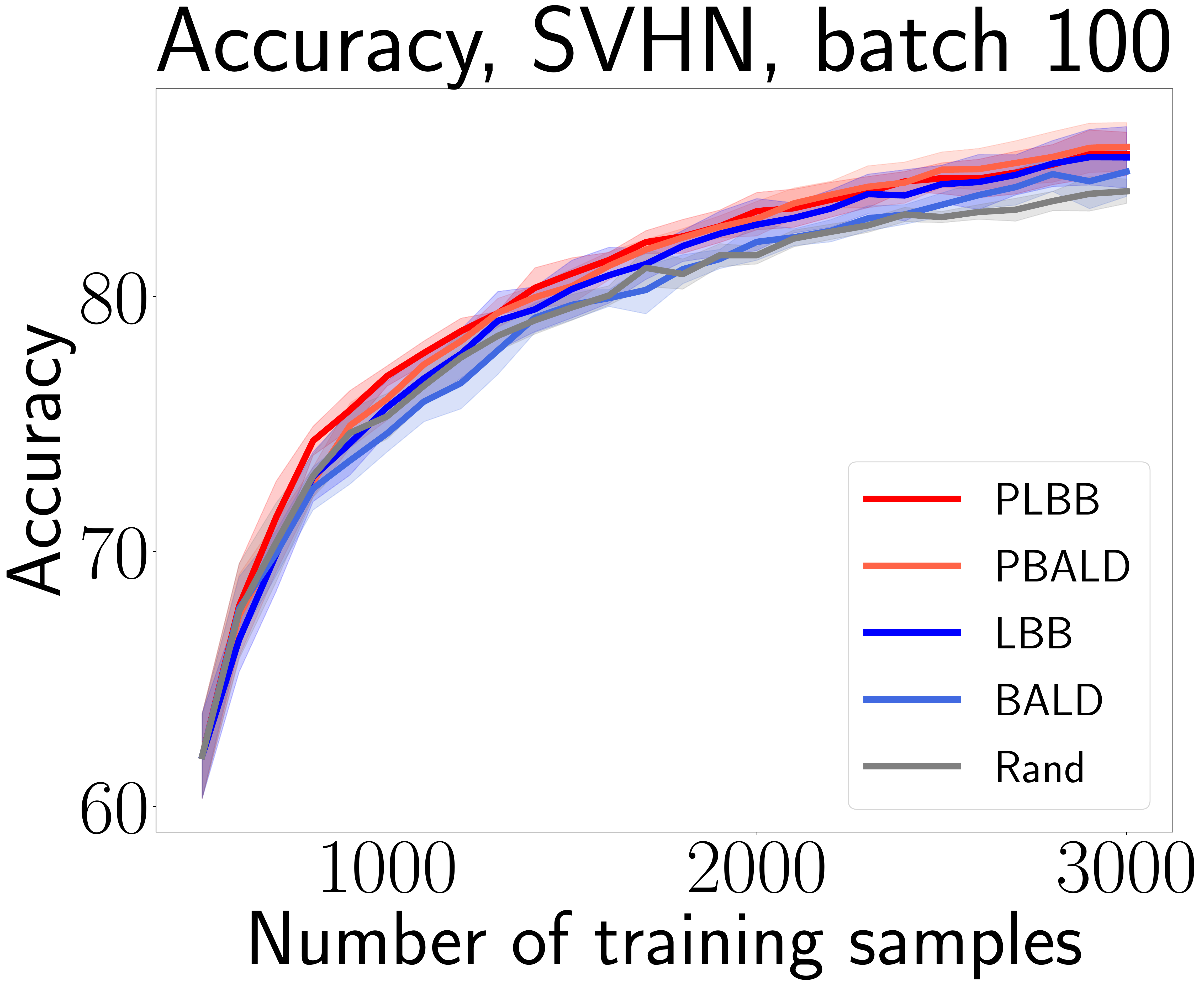}
      \caption{} \label{fig:ens_svhn_batch100}
      \end{subfigure}
      \caption{Performance comparison of AL algorithms on deep ensembles. Datasets: (a)~RCIFAR-10. (b)~RCIFAR-100. (c)~SVHN. }
    \end{figure*}

    Regarding the SVHN dataset with a batch size of $100$ samples, see Figure~\ref{fig:ens_svhn_batch100}, Large BatchBALD shows better quality compared to BALD.
    In turn, BALD has a similar performance to the random baseline.
    In the same figure, the PLBB algorithm slightly outperforms PBALD on the first few thousand elements, after which they have similar quality superior to the competitors.
    The results for RCIFAR-10 on a batch size $200$ are shown in Figure~\ref{fig:ens_rcifar10_batch200}.
    Also, on RCIFAR-100 with a larger batch the Large BatchBALD algorithm outperforms the BALD algorithm, see Figure~\ref{fig:ens_rcifar100_batch500}.

\subsection{Additional results for text data}
\label{sec:appendix_texts}    
\begin{figure}[t!]
        \centering
        \includegraphics[width=0.8\linewidth]{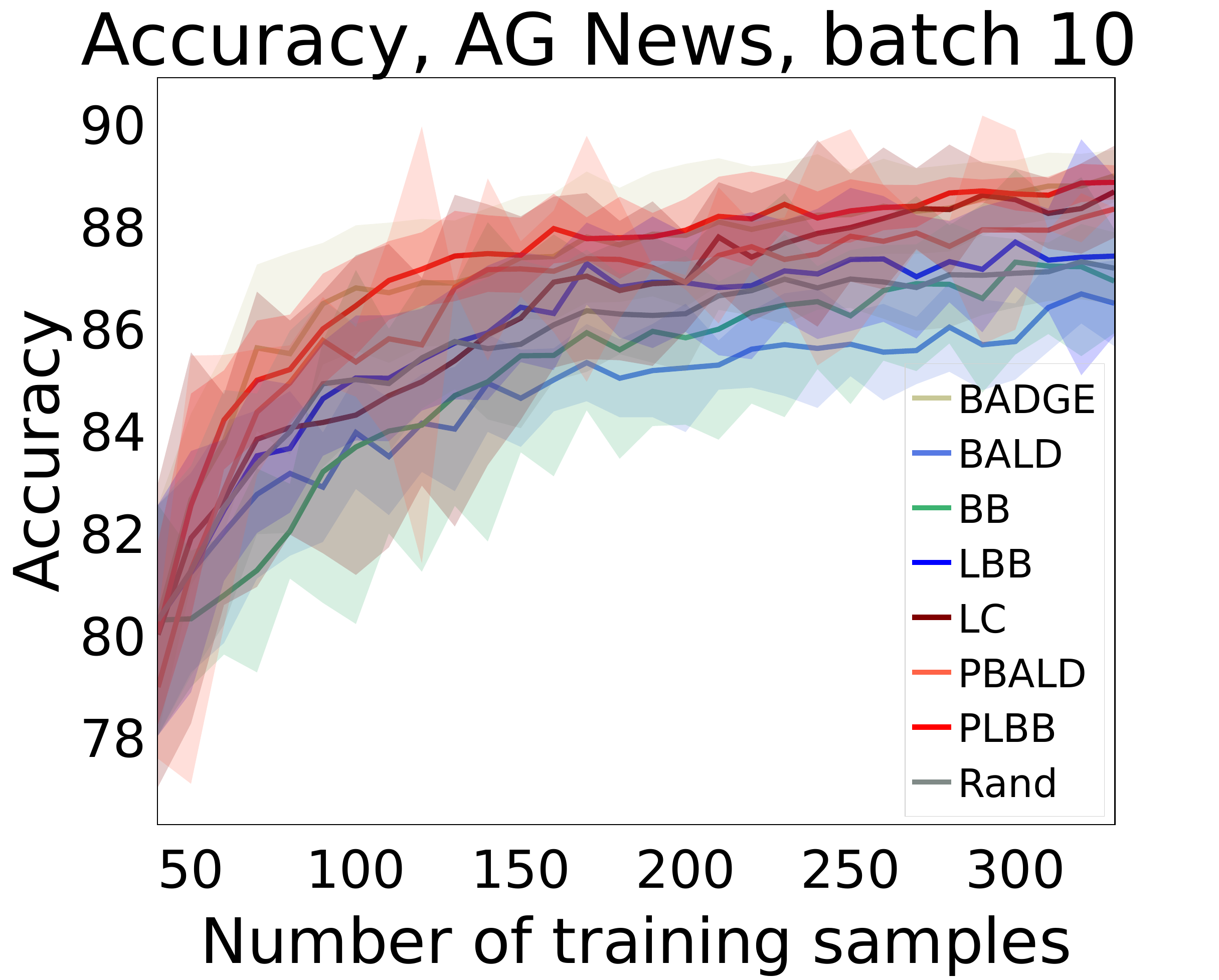}
        \caption{Performance comparison of AL algorithms on AG News dataset with a batch $10$. PLBB and BADGE have comparable top-performance.} \label{fig:ag_small_full}
    \end{figure}

\begin{figure}[t!]
        \centering
        \includegraphics[width=0.8\linewidth]{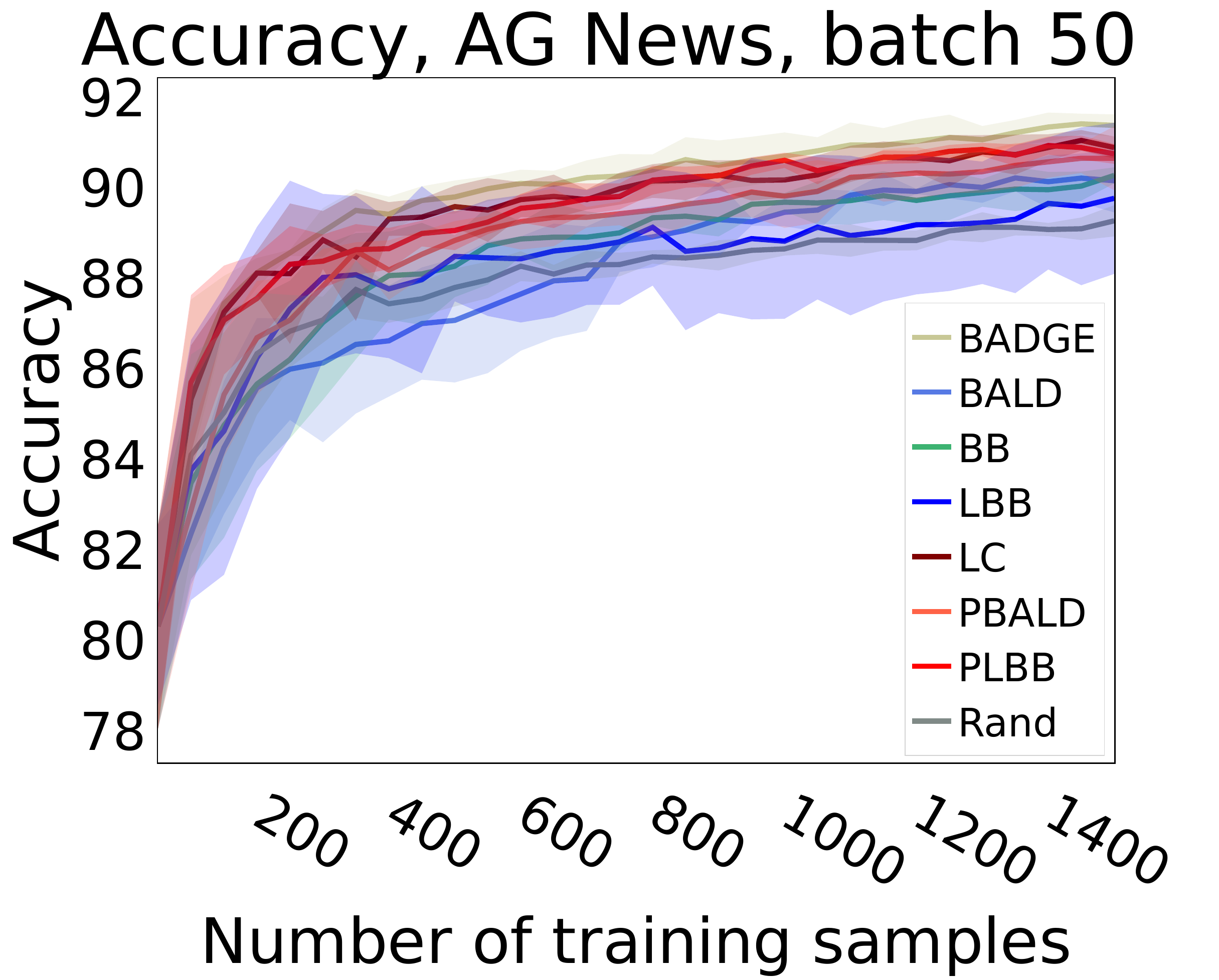}
        \caption{Performance comparison of AL algorithms on AG News dataset with a batch $50$. PLBB and BADGE are two top-performers among alternatives.} \label{fig:ag_big_full}
    \end{figure}

    In addition, we provide figures for AG news dataset, collecting all the considered algorithms, including also the Least Confidence (LC) method and the current SOTA algorithm BADGE, in which additional clustering provides diversity of selected samples.
    As we can see, our algorithm shows a top-performance comparable to BADGE for both smaller ($10$) and larger ($50$) batch sizes, see Figure~\ref{fig:ag_small_full} and Figure~\ref{fig:ag_big_full}, respectively.
    At the same time, LBB shows higher quality compared to BALD and BatchBALD on a small batch, while losing the quality advantage when running the algorithm on a larger batch.
    This may be due to some performance saturation without additional randomization, or the difficulty of dealing with a small number of classes in the presence of a large amount of data.

    \captionsetup{belowskip=0pt}
    
    \captionsetup{belowskip=0pt}
    As a generalization, we analyze the effectiveness of the proposed method compared to alternatives using the so-called Dolan-More curves, which we will discuss next.

\subsection{Dolan-More curves}
\label{sec:dolan_more}
    \begin{figure}[t!]
        \centering
        \includegraphics[width=0.9\linewidth]{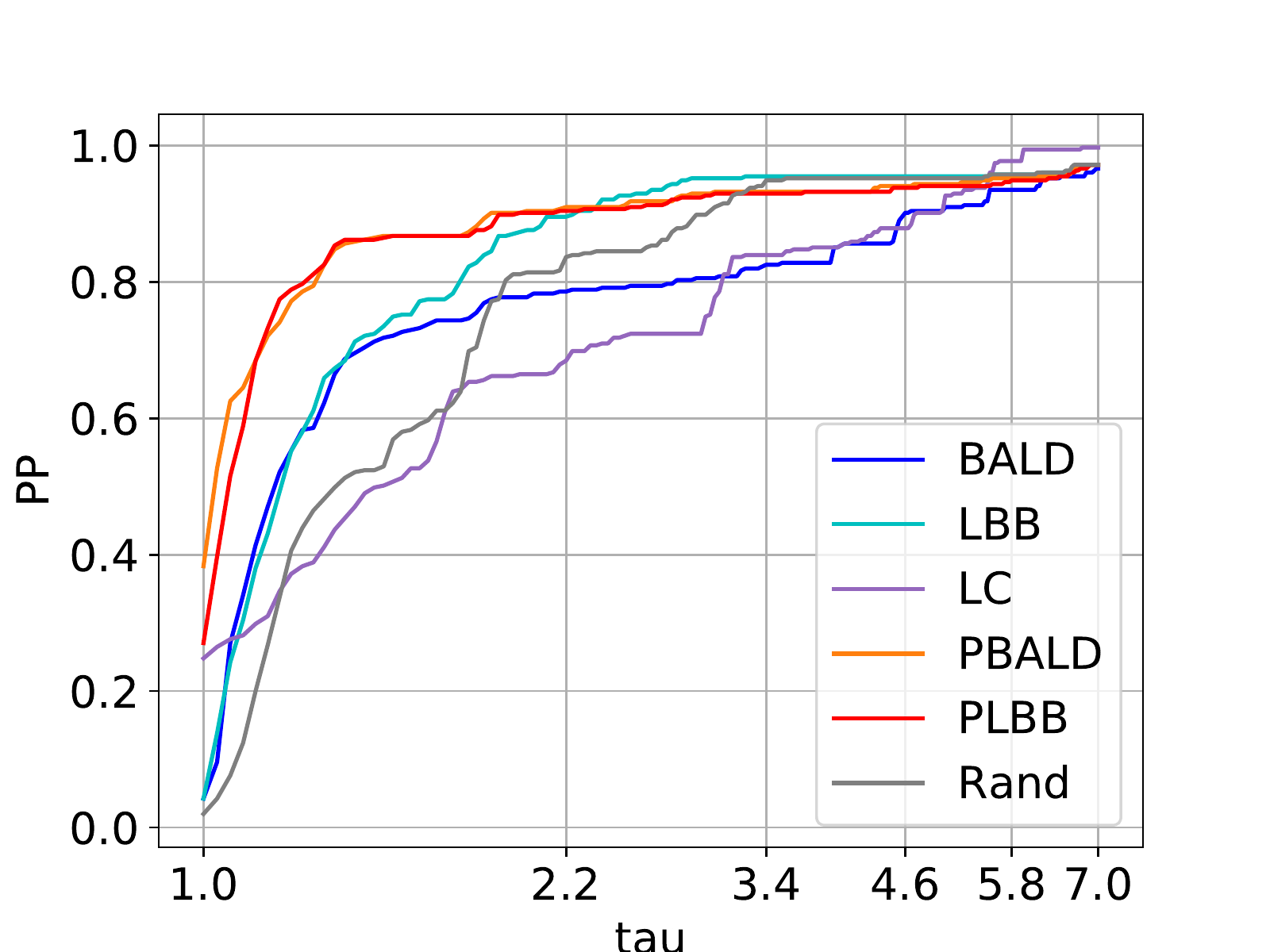}
        \caption{A point on a Dolan-More curve represents the fraction of times when the algorithm is no more than $\tau$ times worse than the best algorithm in terms of error rate (1 - accuracy).
        On the plot six algorithms are compared: BALD, PBALD, LBB, PLBB, Rand and Least Confidence.
        The performance data is aggregated over a number of datasets such as SVHN, CIFAR-100, EMNIST, etc.} \label{fig:dolan_more}
    \end{figure}

    Here, we summarize our results in the form of Dolan-More curves, briefly presenting corresponding methodology and discussing the obtained results.
    
    Let a solver (one of compared algorithms for active learning) be denoted as $s$.
    Let a problem (specific dataset) be denoted as $p$.
    Then for each solver for each dataset, we have the performance measure $t(p, s)$, which is the error rate of the model at the last iteration step of training with the solver $s$ on the dataset $p$.
    Then we introduce the performance ratio:
    \begin{equation*}
        r(p, s) = \frac{t(p, s)}{\min_{s} t(p, s)}.
    \end{equation*}
    The intuition behind the performance ratio is how much the error of a solver is worse than that of the best solver on the given problem.
    In order to get an overall assessment of the performance of the solver, we define:
    \begin{equation*}
        \rho(\tau, s) = \frac{\text{size}\left( p \in P\colon r(p, s) \leq \tau \right)}{|P|},
    \end{equation*}
    where $P$ is the set of all problems.
    The function $\rho(\tau, s)$ can be interpreted as a cumulative distribution function for the performance ratio $r(p, s)$.
    Another name for $\rho(\tau, s)$ is performance profile.
    Ultimately, a plot of the performance profile presents the probability that a solver will be no more than $\tau$ times worse than the best solver as a function of $\tau$.
    
    The Dolan-More curves are presented in the Figure~\ref{fig:dolan_more}.
    The curves are aggregated over $5$ seeds, both MC-dropout and deep ensembles, and all discussed datasets: MNIST, FMNIST, RMNIST, EMNIST, KMNIST, SVHN, CIFAR-10, CIFAR-100, RCIFAR-10, RCIFAR-100 and AG News.
    As we can see, our suggested method LBB is better in terms of performance than BALD.
    Also, their randomized extensions, namely PLBB and PBALD, significantly outperform other algorithms and have a comparable performance to each other.
    Such conclusions on the aggregated results allow us to deduce the successful behavior of the proposed algorithm compared to the alternatives.

\end{document}